\documentclass[10pt,twocolumn,letterpaper]{article}

\usepackage{cvpr}
\usepackage{times}
\usepackage{epsfig}
\usepackage{graphicx}
\usepackage{amsmath}
\usepackage{amssymb}
\usepackage{siunitx}

\usepackage{mycommands}
\usepackage{color}
\usepackage{tabularx}
\usepackage{xpatch}

\usepackage[pagebackref=true,breaklinks=true,letterpaper=true,colorlinks,bookmarks=false]{hyperref}

\cvprfinalcopy %

\def\cvprPaperID{2758} %

\ifcvprfinal\fi
\begin{document}

\title{Detect-to-Retrieve: Efficient Regional Aggregation for Image Search}

\author{Marvin Teichmann\thanks{Both authors contributed equally to this work.}\\
University of Cambridge, UK\\
{\tt\small mttt2@eng.cam.ac.uk}
\and
Andr\'{e} Araujo\footnotemark[1]\quad Menglong Zhu\quad Jack Sim\\
Google AI, USA\\
{\tt\small \{andrearaujo, menglong, jacksim\}@google.com}
}

\maketitle
\makeatletter
\renewcommand\paragraph{\@startsection{paragraph}{4}{\z@}%
                                    {0.5ex \@plus1ex \@minus.2ex}%
                                    {-1em}%
                                    {\normalfont\normalsize\bfseries}}

\makeatother

\begin{abstract}
Retrieving object instances among cluttered scenes efficiently requires compact yet comprehensive regional image representations. Intuitively, object semantics can help build the index that focuses on the most relevant regions. However, due to the lack of bounding-box datasets for objects of interest among retrieval benchmarks, most recent work on regional representations has focused on either uniform or class-agnostic region selection. In this paper, we first fill the void by providing a new dataset of landmark bounding boxes, based on the Google Landmarks dataset, that includes $86k$ images with manually curated boxes from $15k$ unique landmarks. Then, we demonstrate how a trained landmark detector, using our new dataset, can be leveraged to index image regions and improve retrieval accuracy while being much more efficient than existing regional methods. In addition, we introduce a novel regional aggregated selective match kernel (R-ASMK) to effectively combine information from detected regions into an improved holistic image representation. R-ASMK boosts image retrieval accuracy substantially with no dimensionality increase, while even outperforming systems that index image regions independently. Our complete image retrieval system improves upon the previous state-of-the-art by significant margins on the Revisited Oxford and Paris datasets. Code and data available at the project webpage: \url{https://github.com/tensorflow/models/tree/master/research/delf}.
\end{abstract}

\vspace{-15pt}
\section{Introduction}

In this paper, we address the image retrieval problem: given a query image, a system should efficiently retrieve similar images from a database. Image retrieval systems are usually composed of two main stages: (1) \textit{filtering}, where an efficient technique ranks database images according to their similarity with respect to the query; (2) \textit{re-ranking}, where a small number of the most similar database images from the first stage are inspected in more detail and re-ranked.

\begin{figure}[t]
\begin{center}
   \includegraphics[width=1.0\linewidth]{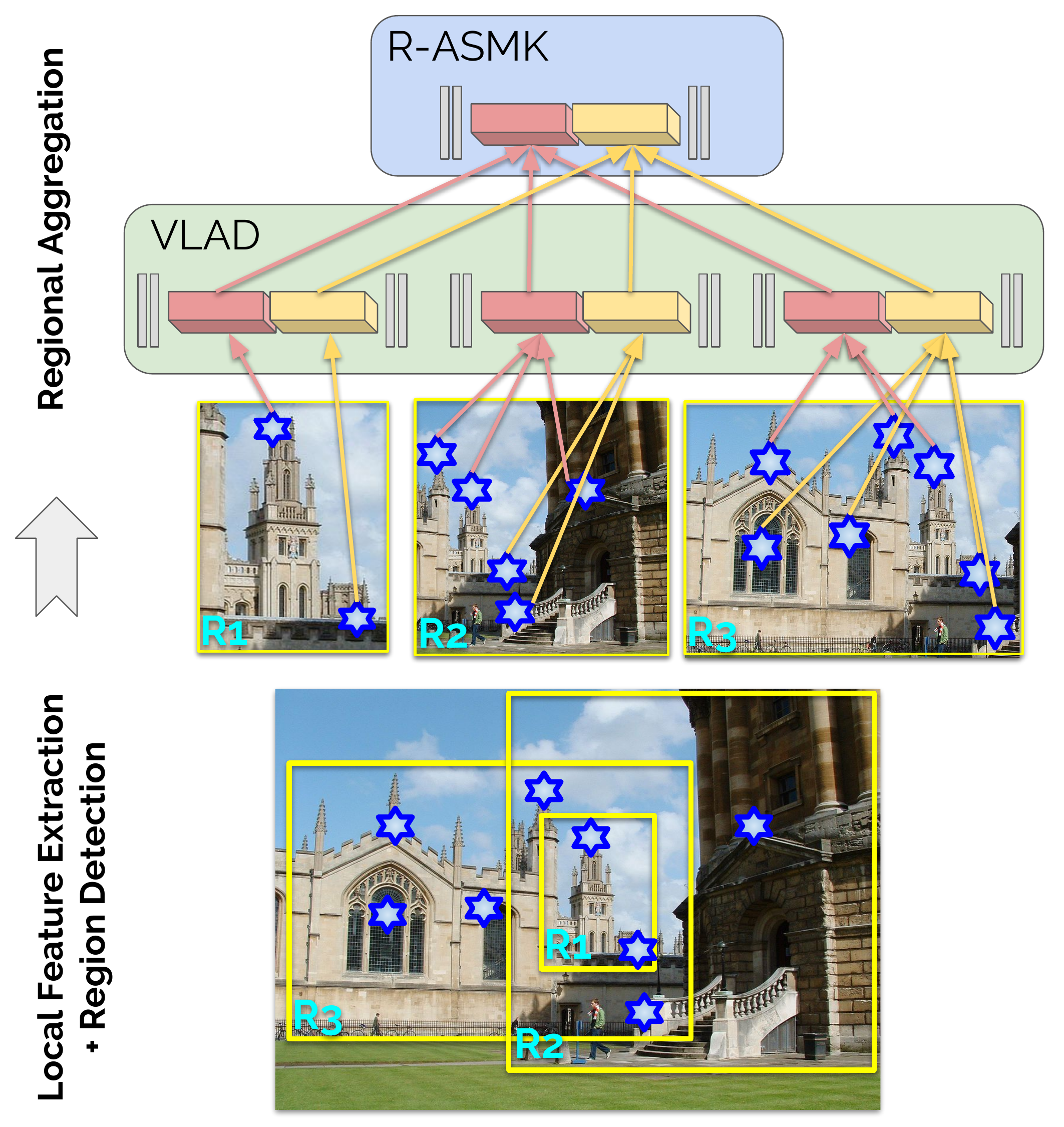}
\end{center}
\vspace{-15pt}
   \caption{
   Overview of our proposed regional aggregation method. Deep local features (stars) and object regions (boxes) are extracted from an image. Regional aggregation proceeds in two steps, using a large codebook of visual words (red and yellow visual words are depicted): first, per-region VLAD description; second, sum pooling and per-visual word normalization. Our final regionally aggregated image representation can be combined to selective match kernels and provide improved image similarity estimation: we refer to this technique as regional aggregated selective match kernels (R-ASMK). It leverages detected regions to improve image retrieval at no dimensionality increase when compared to the original ASMK method \cite{tolias2015image}.
   }
\label{fig:key_fig}
\vspace{0pt}
\end{figure}

Traditionally, hand-crafted local features \cite{Lowe2004,bay2008speeded} were coupled to Bag-of-Words-inspired techniques \cite{Sivic2003, Philbin07, Philbin2008, Jegou2008, jegou2010aggregating, jegou2012aggregating, tolias2015image} to construct high-dimensional representations used in the filtering step. Local feature matching and geometric verification \cite{Philbin07, Philbin2008,avrithis2014hough} (commonly using RANSAC \cite{Fischler1981}) have been used as effective re-ranking strategies.
Recently, several deep learning techniques have been proposed for these two stages. Global image representations based on convolutional neural networks (CNN) can produce compact embeddings to enable fast similarity computation in the filtering step \cite{babenko2014neural,babenko2015iccv,tolias2015particular,arandjelovic2016netvlad,gordo2016deep,radenovic2018fine}. Local image representations can also be extracted using CNNs, suitable to re-ranking via spatial matching and geometric verification \cite{noh2017large,mishkin2018repeatability,mishchuk2017working}.

Today's image retrieval systems tend to fail when relevant objects do not occupy a large enough fraction of database images, typically in cluttered scenes. Often, these objects produce local features that can be used to find local matches against the query in the re-ranking stage. However, such cluttered images usually fail to reach the re-ranking stage, since their initial representation does not lead to high similarity when compared to the query during the filtering stage.
The most common solution to estimate an improved similarity with respect to the query image is to extract and separately store image representations for regions-of-interest in the database, using a fixed regional grid \cite{arandjelovic2013all,razavian2016visual} or a class-agnostic detector \cite{TaoCVPR2014,kim2015predicting}. However, the existing region selection techniques produce a large number of irrelevant regions. In a recent large-scale experimental image retrieval evaluation, Radenovic \etal \cite{radenovic2018revisiting} concluded that such regional search approaches impose too high of a cost in terms of memory and latency, with only small accuracy gains.

\paragraph{Contributions.}
(1) Our first contribution is aimed at improving region selection: we introduce a dataset of manually boxed landmark images, with $86k$ images from $15k$ unique classes, and we show that detectors can be trained for robust landmark localization. (2) Our second contribution is to leverage the trained detector and produce more efficient regional search systems, which improve accuracy for small objects with only a modest increase to the database size -- much more efficiently than previously proposed techniques. (3) In our third contribution, we propose regional aggregated match kernels to leverage selected image regions and produce a discriminative image representation, illustrated in \figref{fig:key_fig}. This new representation outperforms regional search systems significantly, while at the same time being more efficient: only one descriptor needs to be stored per image.
Our image retrieval system outperforms previously published results by $9.3\%$ absolute mean average precision on the Revisited Oxford-Hard dataset, and $1.9\%$ on the Revisited Paris-Hard dataset \cite{radenovic2018revisiting}.

\vspace{-5pt}
\section{Related Work}
\vspace{-5pt}

\paragraph{Datasets.}
To the best of our knowledge, no manually curated datasets of landmark bounding boxes exist.
Gordo \etal \cite{gordo2016deep} use SIFT \cite{Lowe2004} matching to estimate boxes in landmark images. Such boxes are biased towards the feature extraction and matching technique, and may contain localization errors. Their dataset contains $49k$ boxed images, from $586$ landmarks. In comparison, we use human raters to annotate the regions of interest, and produce $86k$ boxed images from $15k$ landmarks. The OpenImages dataset \cite{kuznetsova2018open} contains $9M$ images, annotated with generic object bounding boxes. Some of them may be considered landmarks, for example: buildings, towers, skyscrapers, billboards. However, these classes make for a small fraction of the entire dataset.

\paragraph{Regional search and aggregation.}
Region selection has been explored in image retrieval systems. They have been used with two different purposes: (i) regional search: selected regions are encoded independently in the database, allowing for retrieval of subimages; (ii) regional aggregation: selected regions are used to improve image representations. In the following, we review these two types of approaches.

\textit{Regional search.} Many papers propose to describe regions using VLAD \cite{jegou2010aggregating} or Fisher Vectors \cite{jegou2012aggregating}: Arandjelovic and Zisserman \cite{arandjelovic2013all} use a multi-scale grid to extract $14$ regions per image; Tao \etal \cite{TaoCVPR2014} use Selective Search \cite{UijlingsIJCV2013} with thousands of regions per image; Kim \etal \cite{kim2015predicting} use maximally stable extremal regions (MSER) \cite{matas2004robust}. Razavian \etal \cite{razavian2016visual} use a multi-scale grid with $30$ regions per image, and compute the similarity of two images by taking into account the distances between all region pairs. Iscen \etal \cite{iscen2017efficient, iscen2018fast} leverage multi-scale grids in conjunction with CNN features \cite{radenovic2016cnn}, to enable query expansion via diffusion.
More recently, Radenovic \etal \cite{radenovic2018revisiting} performed a comprehensive evaluation of retrieval techniques and concluded that existing regional search methods may improve recognition accuracy, however at significantly larger memory and complexity costs. In contrast, our Detect-to-Retrieve framework aims at efficient regional search via the use of a custom trained detector.

\textit{Regional aggregation.} 
Tolias \etal \cite{tolias2015particular} leverage the grid structure from \cite{razavian2016visual} to pool pretrained CNN features \cite{krizhevsky2012imagenet,Simonyan15} into compact representations; approximately $20$ regions are selected per image. Radenovic \etal \cite{radenovic2016cnn} build upon \cite{tolias2015particular} by re-training features on a dataset collected in an unsupervised manner.
Gordo \etal \cite{gordo2016deep} train a region proposal network \cite{ren2015faster} from semi-automatic bounding box annotations, to replace the grid from \cite{tolias2015particular}. Hundreds of regions per image are considered in this case.
Our work departs from these papers by using a small set of regions (fewer than $5$ per image), and by formulating regional aggregation as a new match kernel (instead of regional sum-pooling as in \cite{tolias2015particular,gordo2016deep}).

\section{Google Landmark Boxes Dataset}\label{sec:data}

\begin{figure*}
\begin{tabular}{ccccc}
\hspace{-0.1cm} \includegraphics[width=0.175 \textwidth]{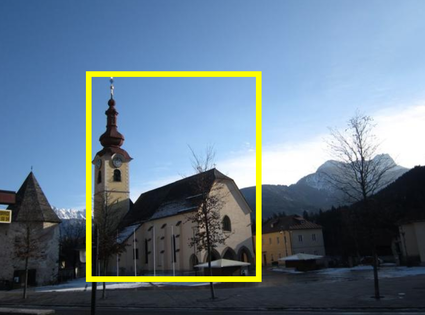}&
\includegraphics[width=0.175 \textwidth]{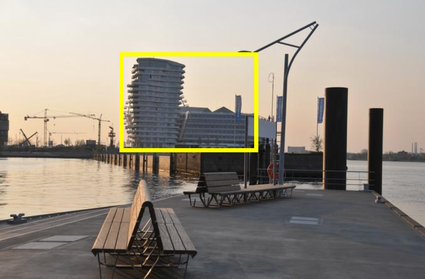}&
\includegraphics[width=0.175 \textwidth]{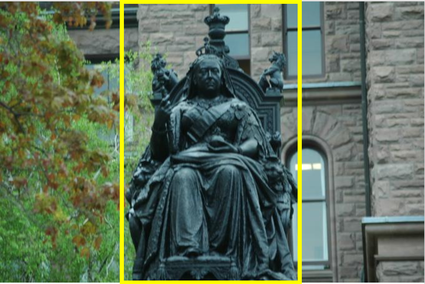}&
\includegraphics[width=0.175 \textwidth]{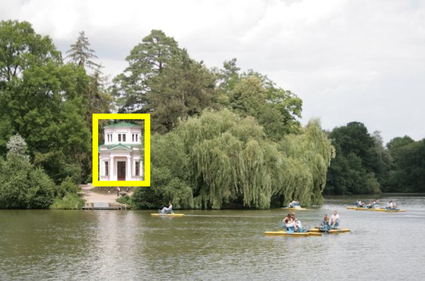}&
\includegraphics[width=0.175 \textwidth]{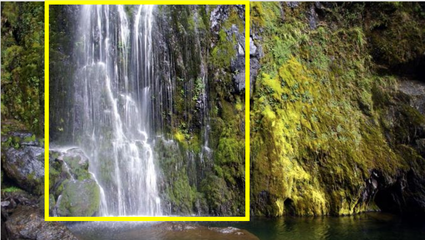}\\

\hspace{-0.1cm} \includegraphics[width=0.175 \textwidth]{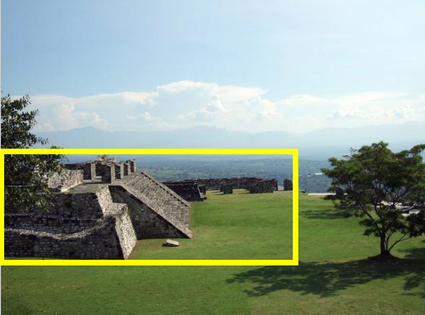}&
\includegraphics[width=0.175 \textwidth]{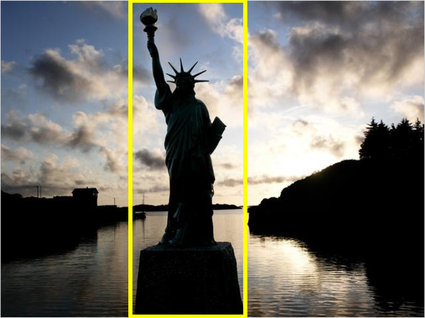} &
\includegraphics[width=0.175 \textwidth]{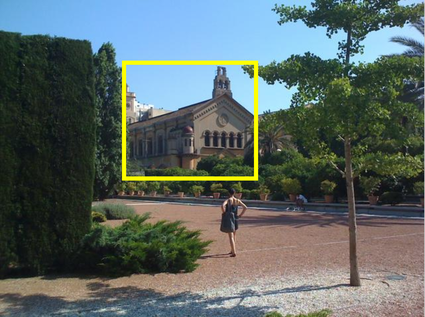}&
\includegraphics[width=0.175 \textwidth]{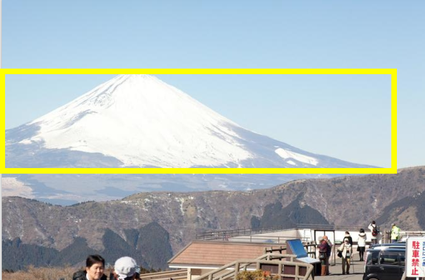}&
\includegraphics[width=0.175 \textwidth]{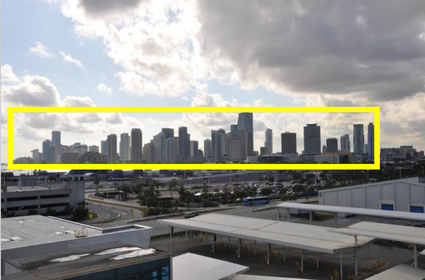}\\

\hspace{-0.1cm} \includegraphics[width=0.175 \textwidth]{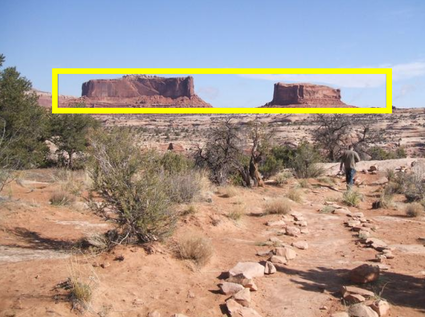}&
\includegraphics[width=0.175 \textwidth]{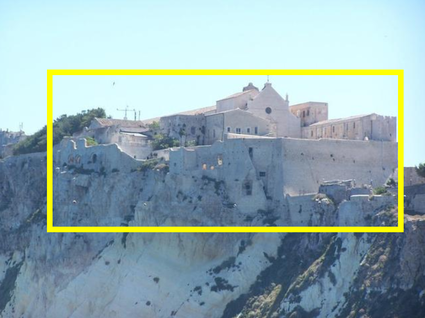} &
\includegraphics[width=0.175 \textwidth]{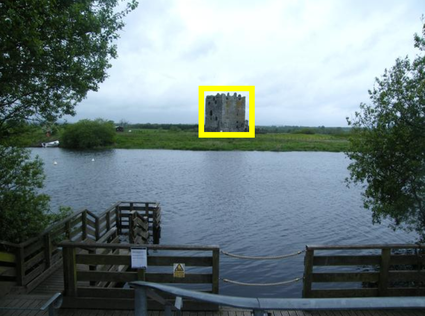}&
\includegraphics[width=0.175 \textwidth]{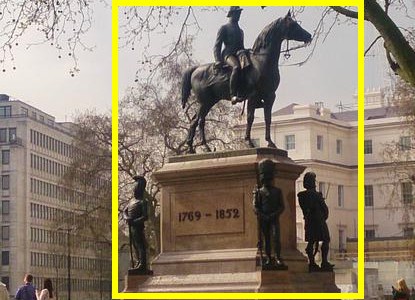}&
\includegraphics[width=0.175 \textwidth]{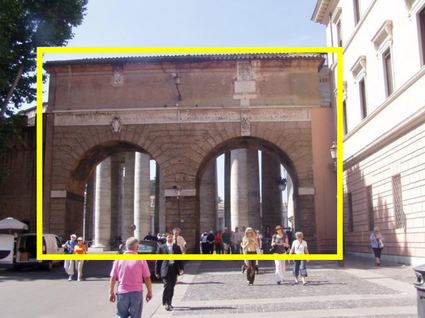}\\

\end{tabular}
\vspace{-10pt}
\caption{Examples of annotated images from our Google Landmark Boxes dataset. A box is drawn around the most prominent landmark depicted in the image. The dataset contains a wide variety of objects, ranging from man-made to natural landmarks.}
\label{fig:dataset_examples}

\end{figure*}

In this section, we introduce our newly collected Google Landmark Boxes dataset, describing the manual annotation process. Our work builds upon the recent Google Landmarks dataset (GLD) \cite{noh2017large}, whose training set contains $1.2M$ images of $15k$ unique landmarks, with a wide variety of objects including buildings, monuments, bridges, statues as well as natural landmarks such as mountains, lakes and waterfalls.

Each image in this dataset is considered to only depict one landmark. In some cases, a landmark may consist of a set of buildings: for example, skylines, which are common in this dataset, are considered as a single landmark.
Since GLD is collected in a semi-automatic manner considering popular touristic locations, it is sometimes ambiguous what the landmark of interest may be. When collecting bounding box annotations, our goal is to capture the most prominent landmark in the image, according to the fact that each image is only assigned one landmark label. Each box should reflect the main object (or set of objects) which is showcased in each dataset image. For this reason, we instructed human operators to draw at most one box per image. 

One of the main challenges in such a fine-grained dataset is the inherent long tail of number of image samples per class. In GLD, some landmarks are associated to several thousands of images, while for about half of the classes only \num{10} or fewer images are provided. Our goal is to represent landmarks in a balanced manner in our new dataset, such that trained detectors are able to localize a wide variety of objects. For this reason, we first separate part of the $1.2M$ training set into a validation set. We randomly select four training and four validation images per landmark. In total, this yields $58k$ and $36k$ boxed images for training and validation, respectively. Note that this means that for about $40\%$ of landmarks, all available images are annotated.

\begin{figure*}
\begin{tabular}{ccccc}

\hspace{-0.cm} 
\includegraphics[width=0.175 \textwidth]{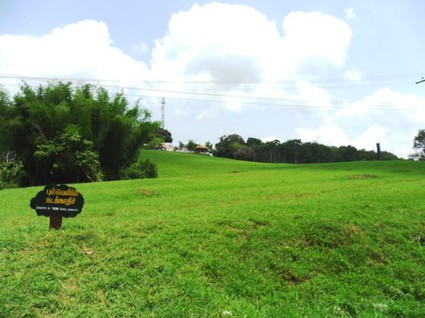}&
\includegraphics[width=0.175 \textwidth]{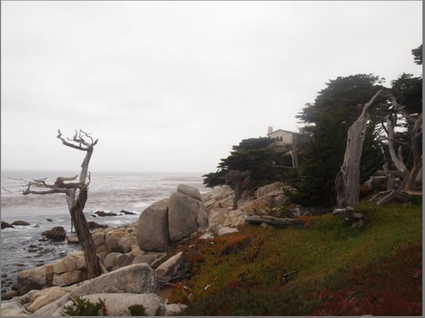}&
\includegraphics[width=0.175 \textwidth]{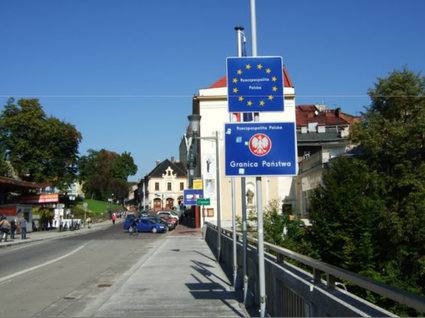}&
\includegraphics[width=0.175 \textwidth]{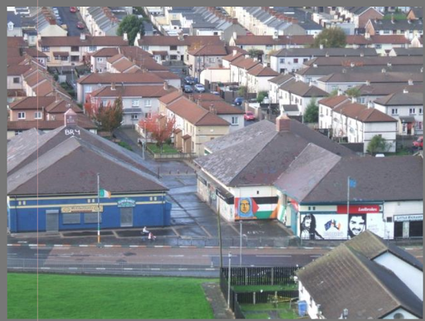}&
\includegraphics[width=0.175  \textwidth]{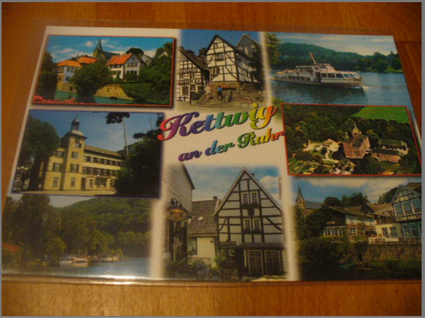}
\end{tabular}
\vspace{-10pt}
\caption{Examples of Google Landmarks dataset images which do not depict a prominent landmark. In such cases (about $8\%$ of images), no boxes were drawn, and the images were not included in the Google Landmark Boxes dataset.}
\label{fig:corner_cases}

\end{figure*}

Examples of annotated images are shown in \figref{fig:dataset_examples}. In some cases, it is not possible to identify a prominent landmark (see \figref{fig:corner_cases}): the landmark of interest may be occluded, or the image may actually show the surroundings of a landmark. We remove such corner cases from our dataset (this applied to about $8\%$ of images which were initially selected), leading to a final dataset with $54k$ and $32k$ boxed images for training and validation, respectively.

\section{Regional Search and Aggregation} \label{sec:method}

We present techniques that enhance image retrieval performance by utilizing bounding boxes predicted by a trained landmark detector. In particular, our approach builds on top of deep local features (DELF) \cite{noh2017large} and aggregated selective match kernels (ASMK) \cite{tolias2015image}, which were recently shown to achieve state-of-the-art performance on a large-scale image retrieval benchmark \cite{radenovic2018revisiting}.

\subsection{Background}

We briefly review the aggregated match kernel framework by Tolias \etal \cite{tolias2015image}.
An image $X$ is described by a set $\mathcal{X} = \{x_1,x_2,\ldots,x_M\}$ containing $M$ local descriptors, each of dimension $D$. A codebook $\mathcal{C}$ comprising $C$ visual words, learned using $k$-means, is used to quantize the descriptors. Denote $\mathcal{X}_c = \{x \in \mathcal{X} : q(x)=c\}$ as the subset of descriptors from $X$ which are assigned to visual word $c$ by the nearest neighbor quantizer $q(x)$.

According to this framework, the similarity between two images $X$ and $Y$, represented by local descriptor sets $\mathcal{X}$ and $\mathcal{Y}$, can be computed as:

\begin{equation}
K(X, Y) = \gamma(\mathcal{X})\gamma(\mathcal{Y}) \sum_{c \in \mathcal{C}} \sigma(\Phi(\mathcal{X}_c)^T \Phi(\mathcal{Y}_c)) 
\label{eq:aggregated_kernels}
\end{equation}

\noindent where $\Phi(\mathcal{X})$ is an aggregated vector representation, $\sigma(.)$ denotes a scalar selectivity function and $\gamma(\mathcal{X})=\left(\sum_c \sigma(\Phi(\mathcal{X}_c)^T \Phi(\mathcal{X}_c))\right)^{-1/2}$ is a normalization factor.
This formulation encompasses popular local feature aggregation techniques, such as Bag-of-Words \cite{Sivic2003}, VLAD \cite{jegou2010aggregating} and ASMK \cite{tolias2015image}.

In particular, for VLAD, $\sigma(u)=u$ and $\Phi(\mathcal{X}_c)$ corresponds to an aggregated residual $V(\mathcal{X}_c)=\sum_{x \in \mathcal{X}_c} x - q(x)$. For ASMK, $\sigma(u)$ corresponds to a thresholded polynomial selectivity function

\begin{equation}
\sigma(u) = 
\begin{cases}\!%
\textrm{sign}(u)|u|^{\alpha}, & \text{if } u > \tau\\
0, &\text{otherwise}
\end{cases}
\end{equation}

\noindent where usually $\alpha=3$ and $\tau=0$; and $\Phi(\mathcal{X}_c)$ corresponds to a normalized aggregated residual $\hat{V}(\mathcal{X}_c) = V(\mathcal{X}_c)/\mynorm{V(\mathcal{X}_c)}$.

\subsection{Regional Search}

In this section, we consider image retrieval systems where regional descriptors are stored independently in the database. Denote the query image as $X$, and the database of $N$ images as $\{Y^{(n)}\}$, $n=1,2,\ldots,N$. We are mainly interested in the experimental configuration where a query contains a well-localized region-of-interest (\ie, the query in practice contains only one region), which is a common setting in image retrieval.
For the $n$-th database image, regions $r_n = 1,\ldots,R_n$ are predicted by a landmark detector, defining the subimages $\{Y^{(n,r_n)}\}$. We denote $Y^{(n,1)}=Y^{(n)}$ as the subimage corresponding to the original image, and always consider it as a valid region. To leverage uncluttered representations, we store aggregated descriptors independently for each subimage, which leads to a total of $\sum_{n=1}^N R_n$ items in the database.

To compute the similarity between the query $X$ and a database image $Y^{(n)}$, we consider max-pooling or average-pooling individual regional similarities, respectively:

\begin{align}
\textrm{sim}_{MAX}(X,Y^{(n)}) &= max_{r=1,\ldots,R_n} K\left(\mathcal{X},\mathcal{Y}^{(n,r)}\right) 
\label{eq:max_pooling_similarity} \\
\textrm{sim}_{AVG}(X,Y^{(n)}) &= \frac{1}{R_n}\sum_{r=1}^{R_n} K\left(\mathcal{X},\mathcal{Y}^{(n,r)}\right)  \label{eq:average_pooling_similarity}
\end{align}

Max-pooling corresponds to assigning a database image's score considering only its highest-scoring subimage. Average pooling aggregates contributions from all subimages. These two variants are compared in \secref{sec:experiments}.

\subsection{Regional Aggregated Match Kernels}

Storing descriptors of each region independently in the database incurs additional cost for both memory and search computation.
In this section, we consider utilizing the detected bounding boxes to instead improve the aggregated representations of database images -- producing discriminative descriptors at no additional cost. We extend the aggregated match kernel framework of Tolias \etal \cite{tolias2015image} to regional aggregated match kernels, as follows.

We start by noting that the average pooling similarity \equref{eq:average_pooling_similarity} can be rewritten as:

\begin{align}
& \textrm{sim}_{AVG}(X,Y^{(n)}) = \nonumber \\
& \gamma(\mathcal{X}) \sum_c \sum_r \frac{\gamma(\mathcal{Y}^{(n,r)})}{R_n} \sigma \left( \Phi(\mathcal{X}_c)^T \Phi(\mathcal{Y}_c^{(n,r)}) \right)
\end{align}

\paragraph{Simple regional aggregation.} For VLAD, this can be further expanded as:
\begin{align}
& \textrm{sim}^{(\textrm{R-VLAD})}(X,Y^{(n)}) \nonumber \\
& = \gamma(\mathcal{X}) \sum_c \sum_r \frac{\gamma(\mathcal{Y}^{(n,r)})}{R_n} V(\mathcal{X}_c)^T V(\mathcal{Y}_c^{(n,r)}) \nonumber \\
& = \sum_c \gamma(\mathcal{X}) V(\mathcal{X}_c)^T \sum_r \frac{\gamma(\mathcal{Y}^{(n,r)})}{R_n} V(\mathcal{Y}_c^{(n,r)}) \label{eq:regional_aggregated_avg_vlad_previous} \\
& = \sum_c V_R(\mathcal{X}_c)^T V_R(\{\mathcal{Y}_c^{(n,r)}\}_r)
\label{eq:regional_aggregated_avg_vlad}
\end{align}

\noindent where we define 

\begin{align}
V_R(\{\mathcal{Y}_c^{(n,r)}\}_r) = \frac{1}{R_n} \sum_r \gamma(\mathcal{Y}^{(n,r)}) V(\mathcal{Y}_c^{(n,r)})
\end{align}

Using this definition, note that $V_R(\mathcal{X}_c)=\gamma(\mathcal{X}) V(\mathcal{X}_c)$.
This derivation indicates that average pooling of regional VLAD similarities can be performed using aggregated regional descriptors and does not require storage of each region's representation separately\footnote{Another way to see that this applies to VLAD kernels is to note that VLAD similarity is computed via a simple inner product, and that the average inner product with a set of vectors equals the inner product with the set average; \ie, for vector $x$ and set $\{y_n\}$, $\frac{1}{N}\sum_n x^Ty_n=x^T\left(\frac{1}{N} \sum_n y_n\right)$.}. We refer to this simple regional aggregated kernel as R-VLAD. 

A similar derivation can be obtained for ASMK in the case where $\sigma(.)$ is the identity function (\ie, no selectivity is applied), by replacing $V(\mathcal{X}_c)$ by $\hat{V}(\mathcal{X}_c)$ in \equref{eq:regional_aggregated_avg_vlad_previous}. A straightforward matching kernel using this idea would apply the selectivity function when comparing the query ASMK representation against this aggregated representation. We refer to this aggregation variant as Naive-R-ASMK.

Both the R-VLAD and Naive-R-ASMK kernels present an important problem when using many detected regions per image and large codebooks. For a given image region, most visual words will not be associated to any local feature, leading to many all-zero residuals for the region. For visual words that correspond to visual patterns observed in only a small number of regions, this will lead to substantially downweighted residuals. We propose to fix this weakness by developing the R-ASMK kernel as follows, inspired by the changes introduced by the original ASMK with respect to VLAD.

\paragraph{R-ASMK.} 
We define the R-ASMK similarity between a query and a database image as:

\begin{align}
& \textrm{sim}^{(\textrm{R-ASMK})}_{}(X,Y^{(n)}) = \nonumber \\
& \sum_c \sigma \left(\hat{V}_R(\mathcal{X}_c)^T \hat{V}_R(\{\mathcal{Y}_c^{(n,r)}\}_r) \right)
\label{eq:regional_aggregated_avg_asmk}
\end{align}

\noindent where $\hat{V}_R(\{\mathcal{Y}_c^{(n,r)}\}_r) = \frac{V_R(\{\mathcal{Y}_c^{(n,r)}\}_r)}{\mynorm{V_R(\{\mathcal{Y}_c^{(n,r)}\}_r)}}$ is the normalized regionally aggregated residual corresponding to visual word $c$.

\paragraph{R-AMK.} The kernels we presented in this section can be regarded as different instantiations of a general regional aggregated match kernel (R-AMK), defined as follows:

\begin{equation}
K_R(X, Y) = \sum_{c \in \mathcal{C}} \sigma \left(\Phi_R(\{\mathcal{X}^{(r)}_c\}_r)^T \Phi_R(\{\mathcal{Y}^{(r)}_c\}_r) \right)
\label{eq:regional_aggregated_kernels}
\end{equation}

\noindent where $\{\mathcal{X}^{(r)}_c\}_r$ denotes the sets of local descriptors quantized to visual word $c$, from each region of $X$. $\Phi_R$ specializes to $V_R$ for R-VLAD, and to $\hat{V}_R$ for R-ASMK. Note that this definition involves regional aggregation for both images, while in this work we focus on the asymmetric case where regional aggregation is applied to the database image only. The asymmetric case is more relevant when the query image is itself a well-localized region-of-interest, which is a common setup in image retrieval benchmarks.

\paragraph{Binarization.} For codebooks with a large number of visual words, the storage cost for such aggregated representations may be prohibitive. Binarization is an effective strategy to allow scalable retrieval in these cases. We adopt a similar binarization strategy as \cite{tolias2015image}, where a binarized version of $\Phi_R$ can be obtained by the elementwise function $b(x)=+1 \, \, \textrm{if}  \, \, x > 0, -1 \, \, \textrm{otherwise}$. We denote the binarized version by a $\star$ superscript (\eg, R-ASMK$^{\star}$ is the binarized version of R-ASMK).

\section{Experiments} \label{sec:experiments}

We present two types of experiments: first, landmark detection, to assess the quality of object detector models trained on the new dataset. Second, we utilize the detected landmarks to enhance image retrieval systems.

\subsection{Landmark Detection} \label{subsec:landmark_detection}

We train two types of detection models on the bounding box data we have collected and described in \secref{sec:data}: a single shot Mobilenet-V2 \cite{sandler2018inverted} based SSD detector \cite{liu2016ssd} and a two stage Resnet-50 \cite{he2015deep} based Faster-RCNN \cite{ren2015faster}. Standard object detection evaluation metric Average Precision (AP) measured at $50\%$ Intersection-over-Union ratio is used during evaluation. Both models reach about $85\%$ AP on the validation set within $500k$ steps ($85.61\%$, $84.37\%$ respectively). The models are trained with publicly available Tensorflow Object Detection API \cite{huang2017speed}. The results indicate that accurate landmark localization can be trained using our dataset. The Mobilenet-V2-SSD variant runs at $27$ms per image, while the Resnet-50-Faster-RCNN runs at $89$ms, both numbers on a TitanX GPU.

\vspace{-3pt}
\subsection{Image Retrieval}
\vspace{-3pt}

We perform regional search and regional aggregation experiments. The following describes the experimental setup.

\paragraph{Datasets.} 
We use the Oxford \cite{Philbin07} and Paris \cite{Philbin2008} datasets, which have recently been revisited to correct annotation mistakes, add new query images and introduce new evaluation protocols \cite{radenovic2018revisiting}; the datasets are referred to as $\mathcal{R}$Oxf and $\mathcal{R}$Par, respectively.
There are $70$ query images for each dataset, with $4993$ ($6322$) database images in the $\mathcal{R}$Oxf ($\mathcal{R}$Par) dataset. We report results on the Medium and Hard setups; for ablations, we focus more specifically on the Hard setup. Performance is measured using mean average precision (mAP) and mean precision at rank $10$ (mP@$10$).
We also perform large-scale experiments using the $\mathcal{R}$1M distractor set \cite{radenovic2018revisiting}, which contains \num[group-separator={,}]{1001001} images.

\begin{figure*}[t]
\centering
\begin{subfigure}{.5\textwidth}
\captionsetup{skip=2pt} %
  \centering
  \includegraphics[width=0.9\linewidth]{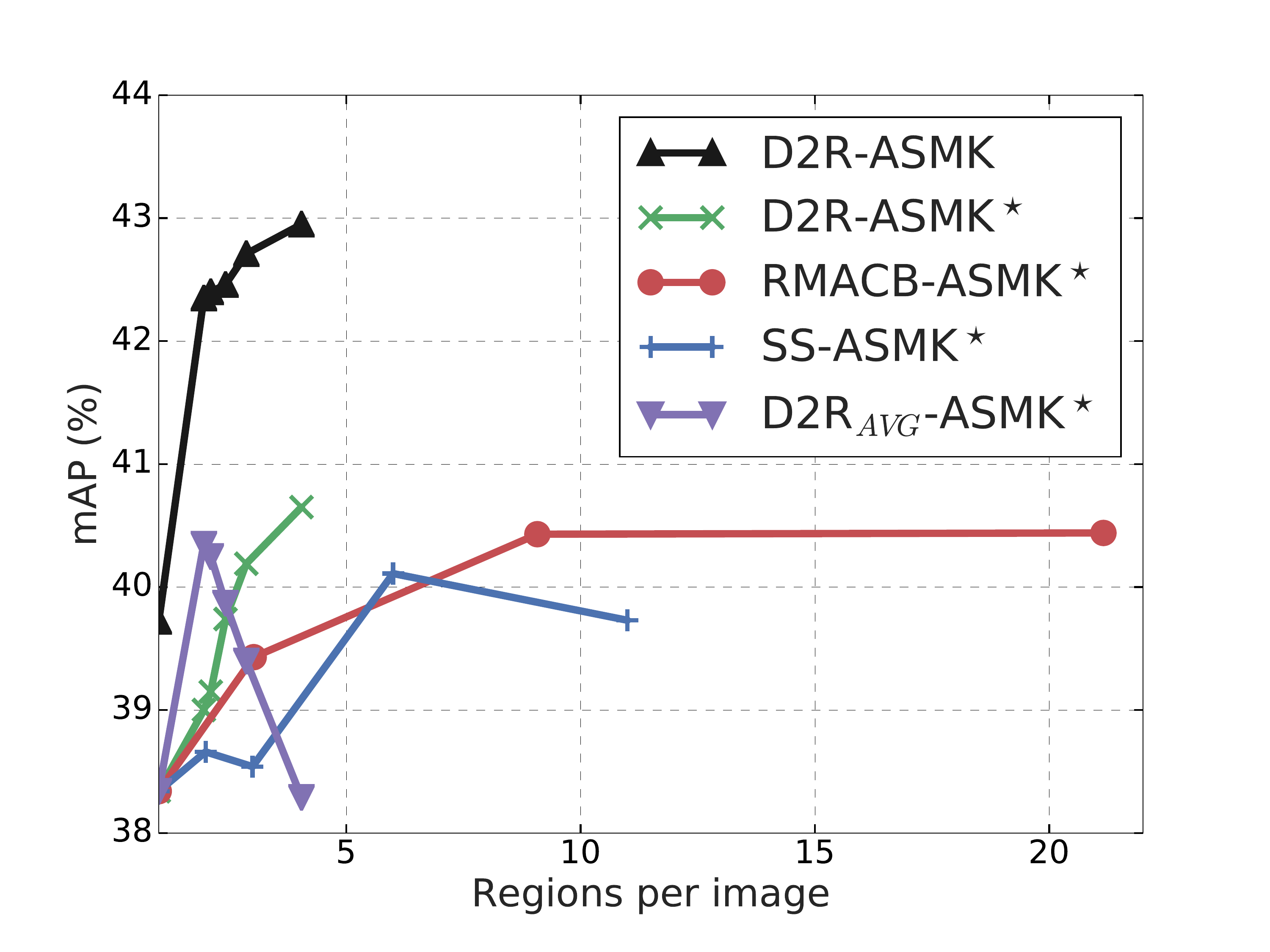}
  \caption{Regional search evaluation.}
  \label{fig:box_comparison}
\end{subfigure}%
\begin{subfigure}{.5\textwidth}
\captionsetup{skip=2pt} %
  \centering
  \includegraphics[width=0.9\linewidth]{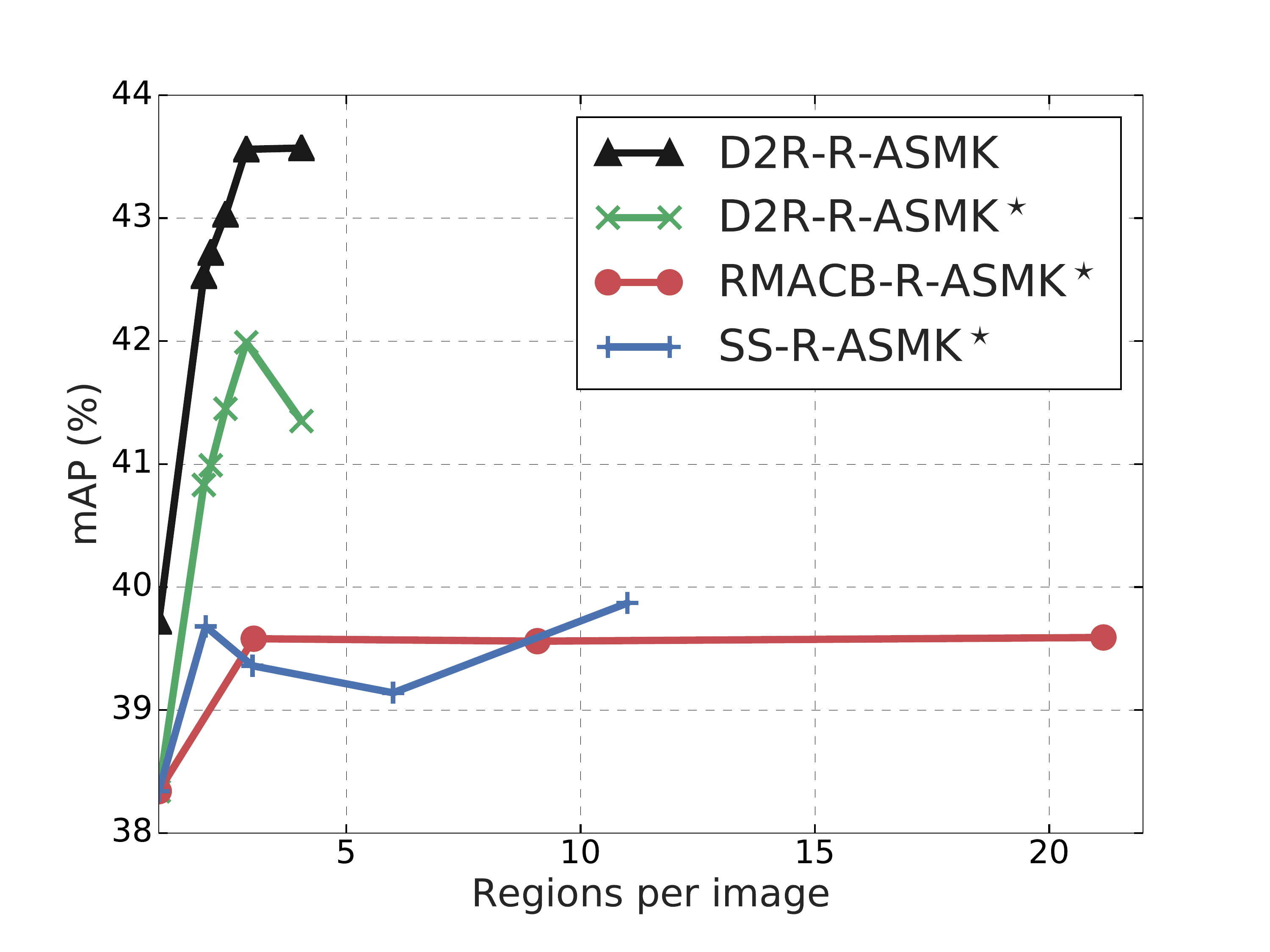}
  \caption{Regional aggregation evaluation.}
  \label{fig:box_comparison_2}
\end{subfigure}
\vspace{2pt}
\caption{Regional search and aggregation evaluations of different image representations, on $\mathcal{R}$Oxf-Hard. (a) Regional search: each regional representation is stored independently in the database, leading to increased memory requirements. Our D2R-ASMK variants achieve significant improvements over the single-image baseline while requiring substantially fewer boxes compared to other region selection approaches. (b) Regional aggregation: each region contributes to the aggregated representation for the entire image. The aggregated descriptor dimensionality is identical to single-image baseline that does not use regions. Our D2R-R-ASMK variants leverage the different landmark regions to compose a strong image representation, which is even more effective than storing each regional representation separately.}
\label{fig:test}
\end{figure*}

\paragraph{Image representation.} We use the following setup in our experiments, except where indicated otherwise. The released DELF model \cite{noh2017large} (pre-trained on the dataset from \cite{gordo2016deep}) is used, with the default configuration (maximum of $1000$ features per region are extracted, with a required minimum attention score of $100$), except that the feature dimensionality is set to $128$ as in previous work \cite{radenovic2018revisiting}.
A $1024$-sized codebook is used when computing aggregated kernels; as common practice, codebooks are trained on $\mathcal{R}$Oxf for retrieval experiments on $\mathcal{R}$Par, and vice-versa.
We focus on improving the core image representations for retrieval, and do not consider query expansion (QE) \cite{chum2007total} techniques such as Hamming QE \cite{tolias2014visual}, $\alpha$ QE \cite{radenovic2018fine} or diffusion \cite{iscen2017efficient, iscen2018fast}; these methods could be incorporated to our system to obtain even stronger retrieval performance.

\paragraph{Region selection techniques.} 
For our Detect-to-Retrieve (D2R) framework, we adopt the trained Faster R-CNN detector described in \secref{subsec:landmark_detection}. We compare against previously proposed region selection techniques for image retrieval: the uniform grid from \cite{razavian2016visual,tolias2015particular} (denoted RMACB, for ``RMAC boxes'') and Selective Search (SS) \cite{UijlingsIJCV2013, TaoCVPR2014}. To vary the number of regions per image, we do as follows: (i) for D2R, we vary the landmark detector threshold; (ii) for RMACB, we sweep the number of levels from $1$ to $3$; (iii) for SS, we select the top $\{1, 2, 5, 10\}$ boxes per image (as in this case there are no confidence scores associated to regions). For all region selection techniques, we add the original image as one of the selected regions.

\begin{table}
\centering
\begin{tabular}{l c c c c c}
\toprule
\multirow{2}{*}{Method} & Det. & \multicolumn{2}{c}{$\mathcal{R}$Oxf-Hard } & \multicolumn{2}{c}{$\mathcal{R}$Par-Hard} \\
 &  Thresh. & mAP & Size & mAP & Size \\
\midrule
 ASMK$^\star$     & ---        & \num{38.3} & \num{1} & \num{54.2} & \num{1} \\ \midrule
  &  \num{0.7} & \num{39.2} & \num{2.1} & \num{56.0} & \num{2.2} \\
 D2R- &  \num{0.5} & \num{39.7} & \num{2.4} & \num{56.2} & \num{2.4} \\
 ASMK$^\star$ &  \num{0.3} & \num{40.2} & \num{2.9} & \num{56.3} & \num{2.9} \\
  &  \num{0.1} & \num{40.7} & \num{4.1} & \num{56.7} & \num{3.9} \\ \midrule
  &  \num{0.7} & \num{41.0} & \num{1} & \num{56.2} & \num{1} \\
  D2R-R- &  \num{0.5} & \num{41.5} & \num{1} & \num{56.2} & \num{1} \\
  ASMK$^\star$ &  \num{0.3} & \num[math-rm=\mathbf]{42.0} & \num{1} & \num{56.3} & \num{1} \\
  &  \num{0.1} & \num{41.4} & \num{1} & \num[math-rm=\mathbf]{56.8} & \num{1} \\ 
\bottomrule
\end{tabular}
\vspace{-8pt}
\caption{Retrieval mAP and relative database size for the different region-based techniques introduced in this work, on the $\mathcal{R}$Oxf-Hard and $\mathcal{R}$Par-Hard datasets, as a function of the landmark detector threshold used for region selection. D2R-ASMK$^\star$ uses max-pooling similarity from \equref{eq:max_pooling_similarity}. The performances of both D2R-ASMK$^\star$ and D2R-R-ASMK$^\star$ tend to improve as the detection threshold decreases (more regions are selected). D2R-R-ASMK$^\star$ outperforms D2R-ASMK$^\star$ consistently, with a smaller memory footprint.}
\label{tab:search_aggregation_comparison}
\end{table}

\paragraph{Implementation details.} We implemented the aggregated kernel framework from scratch in Python/Tensorflow. As a comparison against the reference MATLAB implementation \cite{tolias2015image}, our ASMK$^\star$ with a $1024$-sized codebook and DELF features obtains $37.91\%$ mAP in the $\mathcal{R}$Oxf-Hard dataset, while the reference implementation obtains $37.08\%$. Note that the reference implementation uses a similar configuration as Hamming Embedding (HE) \cite{Jegou2008}, with a projection matrix before binarizarion, residuals computed with respect to the median, and IDF. We did not find consistent improvements using these, so we use the simpler version as described in \secref{sec:method}. Similarly, the reference implementation uses multiple visual word assignments, but our preliminary experiments show improved results using single assignment, making retrieval faster and simpler -- therefore we adopt single assignment in our experiments.
We extended this implementation to support our regional search and aggregation techniques. 

\vspace{-6pt}
\subsubsection{Regional Search}
\vspace{-1pt}

\begin{table*}
\addtolength{\tabcolsep}{-0.15em}
\centering
{\scriptsize
\begin{tabular}{| l | r r | r r | r r | r r | r r | r r | r r | r r |}
\toprule
 \small \multirow{3}{*}{Method}  & \multicolumn{8}{c|}{ \small Medium} & \multicolumn{8}{c|}{ \small Hard} \\[+0.3em] \cline{2-17}
 & \multicolumn{2}{c|}{\large \vphantom{M} \scriptsize $\mathcal{R}$Oxf } & \multicolumn{2}{c|}{\scriptsize $\mathcal{R}$Oxf+$\mathcal{R}$1M} & \multicolumn{2}{c|}{\scriptsize $\mathcal{R}$Par} & \multicolumn{2}{c|}{\scriptsize $\mathcal{R}$Par+$\mathcal{R}$1M} & \multicolumn{2}{c|}{\scriptsize $\mathcal{R}$Oxf} & \multicolumn{2}{c|}{\scriptsize $\mathcal{R}$Oxf+$\mathcal{R}$1M} & \multicolumn{2}{c|}{\scriptsize $\mathcal{R}$Par} & \multicolumn{2}{c|}{\scriptsize $\mathcal{R}$Par+$\mathcal{R}$1M} \\
 & \tiny mAP & \tiny mP@10 & \tiny mAP & \tiny mP@10 & \tiny mAP & \tiny mP@10 & \tiny mAP & \tiny mP@10 & \tiny mAP & \tiny mP@10 & \tiny mAP & \tiny mP@10 & \tiny mAP & \tiny mP@10 & \tiny mAP & \tiny mP@10 \\
\midrule
AlexNet-GeM \cite{radenovic2018fine} &   \num{43.3} & \num{62.1} & \num{24.2} & \num{42.8} & \num{58.0} & \num{91.6}  & \num{29.9} & \num{84.6} &  \num{17.1} & \num{26.2} & \num{9.4} & \num{11.9}  & \num{29.7} & \num{67.6} & \num{8.4} & \num{39.6} \\
VGG16-GeM \cite{radenovic2018fine} &     \num{61.9} & \num{82.7}  & \num{42.6} & \num{68.1} & \num{69.3} & \num{97.9}  & \num{45.4} & \num{94.1} &  \num{33.7} & \num{51.0} & \num{19.0} & \num{29.4} & \num{44.3} & \num{83.7} & \num{19.1} & \num{64.9} \\ 
ResNet101-R-MAC \cite{gordo2016deep} &   \num{60.9} & \num{78.1} & \num{39.3} & \num{62.1} & \num{78.9}& \num{96.9} & \num{54.8} & \num{93.9} &  \num{32.4}& \num{50.0}& \num{12.5} & \num{24.9} & \num{59.4}& \num{86.1} & \num{28.0} & \num{70.0} \\ 
ResNet101-GeM \cite{radenovic2018fine} & \num{64.7} & \num{84.7} & \num{45.2} & \num{71.7} & \num{77.2} & \num{98.1}  & \num{52.3} & \num{95.3} &  \num{38.5} & \num{53.0} & \num{19.9} & \num{34.9} & \num{56.3} & \num{89.1} & \num{24.7} & \num{73.3} \\
ResNet101-GeM$\uparrow$+DSM \cite{simeoni2019local} & \num{65.3} & \num{87.1} & \num{47.6} & \num{76.4} & \num{77.4} & \num{99.1}  & \num{52.8} & \num{96.7} &  \num{39.2} & \num{55.3} & \num{23.2} & \num{37.9} & \num{56.2} & \num{89.9} & \num{25.0} & \num{74.6} \\
HesAff-rSIFT-ASMK$^\star$ \cite{tolias2015image} & \num{60.4} & \num{85.6} & \num{45.0} & \num{76.0} & \num{61.2} & \num{97.9}  & \num{42.0} & \num{95.3} &  \num{36.4} & \num{56.7} & \num{25.7} & \num{42.1} & \num{34.5} & \num{80.6} & \num{16.5} & \num{63.4} \\
HesAff-rSIFT-ASMK$^\star$+SP \cite{tolias2015image} &  \num{60.6} & \num{86.1} & \num{46.8} & \num{79.6} & \num{61.4} & \num{97.9}  & \num{42.3} & \num{95.3} &  \num{36.7} & \num{57.0} & \num{26.9} & \num{45.3} & \num{35.0} & \num{81.7} & \num{16.8} & \num{65.3} \\ 
HesAff-HardNet-ASMK$^\star$+SP \cite{mishkin2018repeatability} &  \num{65.6} & \num{90.2} & -- & -- & \num{65.2} & \num{98.9} & -- & -- &  \num{41.1} & \num{59.7} & -- & -- & \num{38.5} & \num{87.9} & -- & -- \\ 
DELF-ASMK$^\star$+SP \cite{noh2017large,radenovic2018revisiting} & \num{67.8} & \num{87.9} & \num{53.8} & \num{81.1}  & \num{76.9} & \num{99.3} & \num{57.3} & \num{98.3}  & \num{43.1} & \num{62.4} & \num{31.2} & \num{50.7} & \num{55.4} & \num[math-rm=\mathbf]{93.4} & \num{26.4} & \num{75.7} \\ 
\midrule
DELF-ASMK$^\star$ (reimpl.)               & \num{65.7} & \num{87.9} & -- & -- & \num{77.1} & \num{98.7} & -- & -- &  \num{41.0} & \num{57.9} & -- & -- & \num{54.6} & \num{90.9} & -- & -- \\
DELF-D2R-R-ASMK$^\star$ (ours)            & \num{69.9} & \num{89.0} & -- & -- & \num{78.7} & \num{99.0} & -- & -- &  \num{45.6} & \num{61.9} &  -- & --  & \num{57.7} & \num{93.0} & -- & -- \\
\multicolumn{1}{|c|}{--- DELF-GLD (ours)} & \num{73.3} & \num{90.0} & \num{61.0} & \num{84.6}  & \num[math-rm=\mathbf]{80.7} & \num{99.1} & \num[math-rm=\mathbf]{60.2} & \num{97.9} &  \num{47.6} & \num{64.3} & \num{33.6} & \num{53.7} & \num[math-rm=\mathbf]{61.3} & \num[math-rm=\mathbf]{93.4} & \num[math-rm=\mathbf]{29.9} & \num{82.4} \\
\midrule
DELF-ASMK$^\star$+SP (reimpl.)             & \num{68.9} & \num{90.9} & -- & -- & \num{76.6} & \num{98.7} & -- & -- & \num{46.6} & \num{66.7} & -- & -- & \num{52.2} & \num{87.6} & -- & -- \\
DELF-D2R-R-ASMK$^\star$+SP (ours)          & \num{71.9} & \num{91.3} & -- & -- & \num{78.0} & \num[math-rm=\mathbf]{99.4} & -- & -- & \num{48.5} & \num{66.7} & -- & -- & \num{54.0} & \num{87.6} & -- & -- \\
\multicolumn{1}{|c|}{--- DELF-GLD (ours)}  & \num[math-rm=\mathbf]{76.0}  & \num[math-rm=\mathbf]{93.4}  & \num[math-rm=\mathbf]{64.0} & \num[math-rm=\mathbf]{87.7} & \num{80.2} & \num{99.1} & \num{59.7} & \num[math-rm=\mathbf]{99.0} & \num[math-rm=\mathbf]{52.4} & \num[math-rm=\mathbf]{70.9} & \num[math-rm=\mathbf]{38.1} & \num[math-rm=\mathbf]{61.3} & \num{58.6} & \num{91.0} & \num{29.4} & \num[math-rm=\mathbf]{83.9} \\
\bottomrule
\end{tabular}
}
\vspace{-5pt}
\caption{Comparison of proposed techniques against state-of-the-art methods, on the $\mathcal{R}$Oxford ($\mathcal{R}$Oxf) and $\mathcal{R}$Paris ($\mathcal{R}$Par) datasets (and their large-scale extensions $\mathcal{R}$Oxf+$\mathcal{R}$1M and $\mathcal{R}$Par+$\mathcal{R}$1M), with Medium and Hard evaluation protocols. Previously published results are presented in the first block of rows. The second and third block of rows present our experimental results, considering systems without and with spatial verification (SP), respectively. In this experiment, we use codebooks with $65k$ visual words, to make our results comparable to previous work \cite{radenovic2018revisiting}. DELF-GLD indicates a version of DELF which we re-trained on the Google Landmarks dataset. Our methods achieve equal or improved performance for all evaluation protocols, datasets and metrics.}
\label{tab:sota}
\end{table*}

We compare aggregated match kernels, region selection techniques and similarity computation methods on the $\mathcal{R}$Oxf-Hard dataset. When performing regional search, multiple regions are selected per image and stored independently in the database, leading to increased memory cost.
\figref{fig:box_comparison} presents results for ASMK variants, where all techniques use max-pooling similarity from \equref{eq:max_pooling_similarity}, except for D2R$_{AVG}$-ASMK$^\star$, which uses average-pooling similarity from \equref{eq:average_pooling_similarity}.
Combining our proposed D2R regions with ASMK enhances mAP by $3.23\%$ when using an average of $4.05$ regions per image.

We compare the different region selection approaches using ASMK$^\star$. Our D2R-ASMK$^\star$ achieves $40.65\%$ mAP when using $4.05$ regions per image, improvement of $2.31\%$ over the single-image ASMK$^\star$ baseline.
Other region selection approaches improve retrieval accuracy, but with significantly larger memory requirements. RMACB-ASMK$^\star$ requires $9.08$ regions/image to achieve  $40.43\%$ mAP (this is $0.22\%$ mAP below the previously mentioned D2R-ASMK$^\star$ operating point, despite requiring $2.24\times$ more memory). SS-ASMK$^\star$ benefits from some regions, while performance decreases when a large number of regions are selected, since many of those regions are irrelevant.

Average pooling of individual regional similarities improves upon the single-image baseline significantly, at low overhead memory requirements: D2R$_{AVG}$-ASMK$^\star$ achieves $40.35\%$ mAP with only $1.96\times$ storage cost. Note that in this case performance drops significantly as more regions are added, since irrelevant regional similarities are added to the final image similarity.
We also experimented with a D2R-VLAD representation: mAP improves from $30.17\%$ (single-image) to $33.87\%$ ($2.87$ regions/image).

\tabref{tab:search_aggregation_comparison} further presents D2R-ASMK$^\star$ results on the $\mathcal{R}$Par-Hard dataset. Regional search enables $2.5\%$ mAP improvement at $3.9$ regions/image.
Note that our D2R approach is effective even if the landmarks in the Google Landmark Boxes dataset present much larger variability than the landmarks encountered in the $\mathcal{R}$Oxf/$\mathcal{R}$Par datasets.

\begin{figure*}[t]
\begin{center}
   \includegraphics[width=1.0\linewidth]{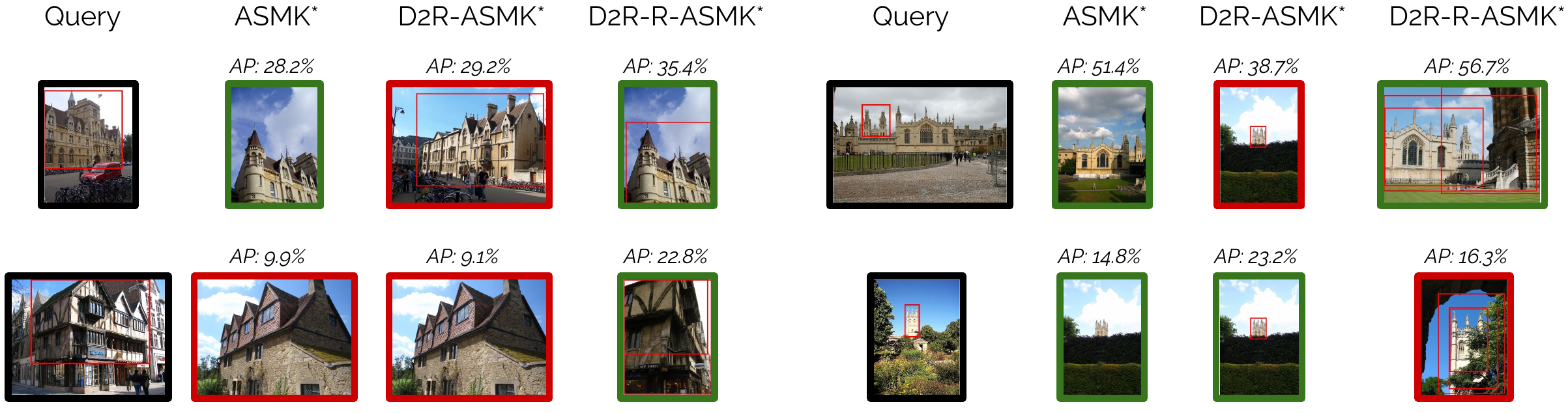}
\end{center}
\vspace{-15pt}
   \caption{Qualitative results for ASMK$^\star$ (baseline single-image method), D2R-ASMK$^\star$ (regional search) and D2R-R-ASMK$^\star$ (regional aggregation) on $\mathcal{R}$Oxf-Hard. Four queries are presented, with their regions-of-interest highlighted. For each method, we show the first ranked image where the methods disagree. Red borders indicate incorrect results, and green borders indicate correct results. For D2R-ASMK$^\star$, we box the region used for the result (or leave unboxed if the region corresponds to the entire image). For D2R-R-ASMK$^\star$, we box all regions used for aggregation. We also present average precision (AP) for each method and query.}
\label{fig:qualitative}
\vspace{0pt}
\end{figure*}

\vspace{-6pt}
\subsubsection{Regional Aggregated Match Kernels}
\vspace{-1pt}

In this section, we evaluate the proposed regional aggregated match kernels. In this experiment, region selection is used to produce an improved image representation, with no increase in the aggregated descriptor dimensionality. \figref{fig:box_comparison_2} compares different aggregation methods and region selection approaches, on the $\mathcal{R}$Oxf-Hard dataset. Both our proposed D2R-R-ASMK and D2R-R-ASMK$^\star$ variants achieve substantial improvements compared to their baselines which do not use boxes for aggregation: $3.85\%$ and $3.65\%$ absolute mAP improvements, respectively. We also compare our D2R approach against other region selection methods. RMACB and SS improve upon the baseline, however with limited gain of at most $1.5\%$ mAP.

More interestingly, our proposed kernels outperform even the regional search configuration where each region is indexed separately in the database. \tabref{tab:search_aggregation_comparison} compiles experimental results on $\mathcal{R}$Oxf-Hard and $\mathcal{R}$Par-Hard. Our D2R-R-ASMK$^\star$ method outperforms the best regional search variant on both datasets, respectively by $1.3\%$ and $0.1\%$ absolute mAP, with relative storage savings of $4.1\times$ and $3.9\times$.

In another ablation experiment, we assess the performance of simpler regional aggregation methods: R-VLAD and Naive-R-ASMK. We use the trained detector to select regions. For R-VLAD, mAP on $\mathcal{R}$Oxf improves from $30.17\%$ (single-image) to $30.91\%$ when using $2.4$ regions per image, but degrades quickly as more regions are considered.
In particular, when setting a very low detection threshold ($0.01$) to obtain $10.2$ regions per image, performance degenerates to $16.46\%$ mAP -- this agrees with the intuition that a large number of regions is detrimental to R-VLAD.
For Naive-R-ASMK, no improvement is obtained when detected regions are used: mAP drops from $39.72\%$ to $31.42\%$ when $1.96$ regions per image are used, and similarly degenerates to $9.2\%$ when using $10.2$ regions per image.
In comparison, using the same detection threshold of $0.01$, R-ASMK$^\star$ obtains $41.6\%$ mAP, \ie, performance is high even if using a large number of regions, due to the improved aggregation technique.

\vspace{-6pt}
\subsubsection{Comparison Against State-of-the-Art}
\vspace{-2pt}

We compare our D2R-R-ASMK$^\star$ technique against state-of-the-art image retrieval systems. To make our system comparable with previously published results \cite{radenovic2018revisiting}, for this experiment we use a codebook with $65k$ visual words. We also further experiment with re-training the DELF local feature on the Google Landmarks dataset (denoted as DELF-GLD). Spatial verification (SP) is used to re-rank the top \num{100} database images (we use RANSAC with an Affine model).

Table \ref{tab:sota} presents experimental results on $\mathcal{R}$Oxf and $\mathcal{R}$Par, using the Medium and Hard protocols, also including the large-scale setup with $\mathcal{R}$1M. Our proposed D2R-R-ASMK$^\star$ representation by itself, without spatial verification, already improves mAP when comparing against all previously published results. SP further boosts performance by about $3\%$ mAP on $\mathcal{R}$Oxf; surprisingly, it slightly degrades performance on the $\mathcal{R}$Par dataset. Re-training DELF on GLD improves performance by around $4\%$. Our best results improve upon the previous state-of-the-art by $8.2\%$ mAP on $\mathcal{R}$Oxf-Medium, $1.8\%$ mAP on $\mathcal{R}$Par-Medium, $9.3\%$ mAP on $\mathcal{R}$Oxf-Hard and $1.9\%$ in $\mathcal{R}$Par-Hard (with similar gains in the large-scale setup).

\paragraph{Memory.} Our DELF-D2R-R-ASMK$^\star$ descriptors have the exact same dimensionality as DELF-ASMK$^\star$. However, DELF-ASMK$^\star$ is sparser and consumes less memory in practice: $10.3$GB, compared to $27.6$GB for DELF-D2R-R-ASMK$^\star$, in the large-scale $\mathcal{R}$Oxf+$\mathcal{R}$1M dataset. This is still much less than other local feature based approaches; \eg HesAff-rSIFT-ASMK$^\star$ requires $62$GB \cite{radenovic2018revisiting}, and HesAffNet-HardNet++-ASMK$^\star$ \cite{mishkin2018repeatability} requires approximately $86.8$GB.

\vspace{-6pt}
\subsubsection{Discussion}
\vspace{-2pt}

Our experiments demonstrate that selecting relevant image regions can help boost image retrieval performance significantly. In our regional aggregation method, the detected regions allow for effective re-weighting of local feature contributions, emphasizing relevant visual patterns in the final image representation.
Note, however, that it is crucial to perform \textbf{both} region selection \textbf{and} regional aggregation in a suitable manner. If the selected regions are not relevant to the objects of interest, regional aggregation cannot be very effective, as shown in \figref{fig:box_comparison_2}. Also, our experiments with naive versions of regional aggregation indicate that the aggregation needs to be performed in the right way: this is related to the poor R-VLAD and Naive-R-ASMK results.

It may initially seem unintuitive that the regional search method underperforms when compared to our regional aggregation technique. However, this can be understood by observing some retrieval result patterns, which are presented in \figref{fig:qualitative}. The addition of separate regional representations to the database may help retrieval of relevant small objects in cluttered scenes, as illustrated with the successful bottom-right D2R-ASMK$^\star$ retrieved image. However, it also increases the chances of finding localized regions which are similar but do not correspond to the same landmark, as illustrated with the top two cases.

Regional aggregation, on the other hand, can help retrieval by re-balancing the visual information presented in an image. The top-right D2R-R-ASMK$^\star$ result shows a database image where the detected boxes do not precisely cover the query object; instead, several selected regions cover it, and consequently its features are boosted. A similar case is illustrated in the bottom-left example, where the main detected region in the database image does not cover the object of interest entirely. The features inside the main box are boosted but those outside are also used, generating a more suitable representation for image retrieval.

\vspace{-5pt}
\section{Conclusions}
\vspace{-1pt}
In this paper, we present an efficient regional aggregation method for image retrieval. We first introduce a dataset of landmark bounding boxes, and show that landmark detectors can be trained and leveraged for extracting regional representations. Regional search using our detectors not only provides superior retrieval performance but also much better efficiency than existing regional methods. In addition, we propose a novel regional aggregated match kernel framework that further boosts the retrieval accuracy. Our full system achieves state-of-the-art performance by a large margin on two image retrieval datasets.

\section*{Appendix A. Discussion: Detection Helps Finding Relevant Features}

In this section, we analyze the detector's ability to focus on relevant landmarks by empirically estimating the proportion of relevant local features located within or without predicted bounding boxes. We extract and match local features for image pairs that are known to depict the same landmark. A local feature is declared to be relevant if it is an inlier to a high-confidence estimated geometric transformation.

\begin{figure}[t]
\begin{center}
  \includegraphics[width=1\linewidth]{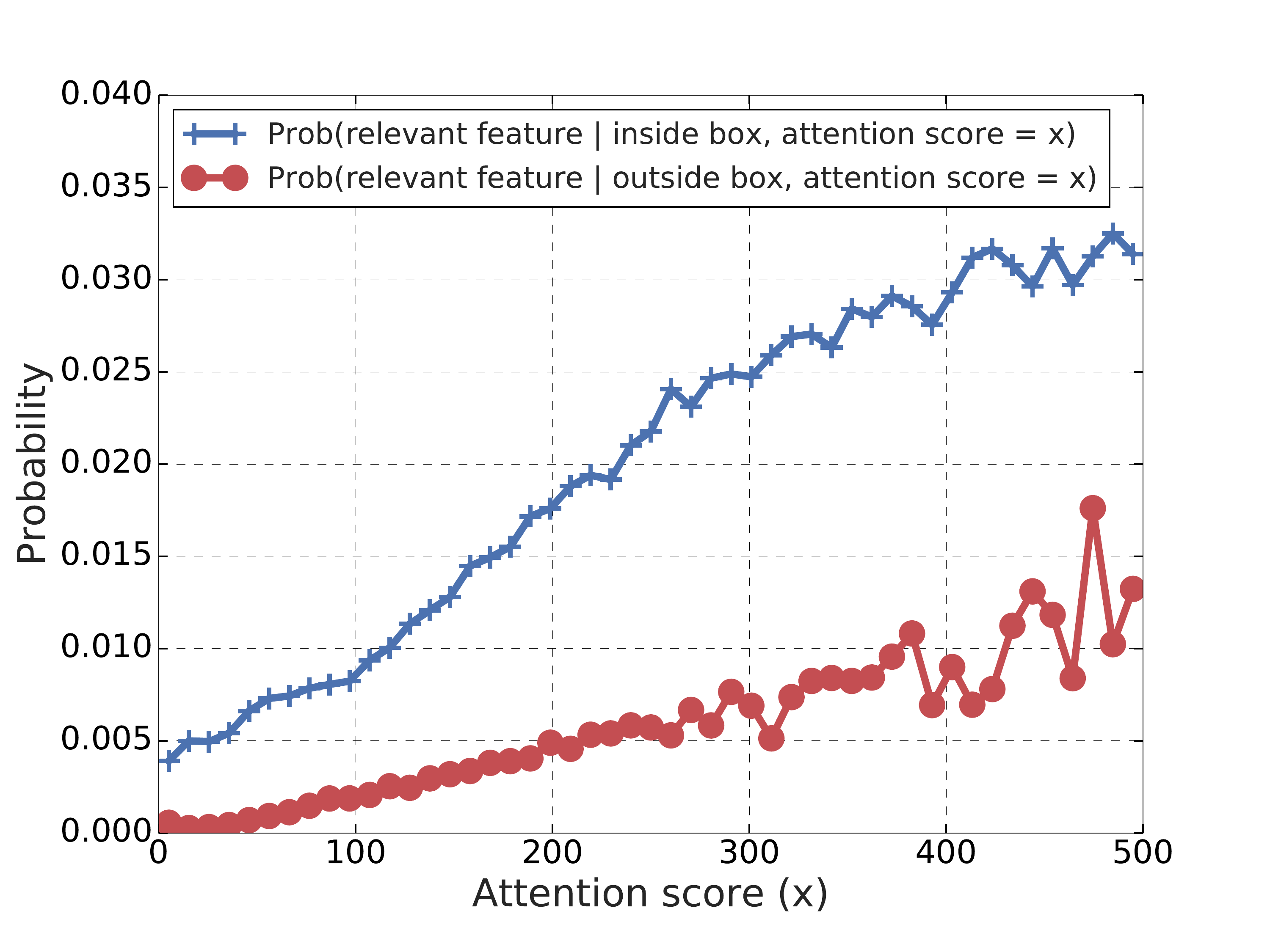}
\end{center}
\vspace{-15pt}
  \caption{Relevance probability of a DELF local feature, as a function of its attention score. The blue curve denotes features inside predicted bounding boxes, while the red curve denotes features outside them. The detected boxes provide valuable information that can be used to improve image representations for retrieval tasks.}
\label{fig:plot_matching}
\vspace{0pt}
\end{figure}

More specifically, we use DELF local features \cite{noh2017large} and a Faster-RCNN \cite{ren2015faster} landmark detector trained on our new dataset. $10k$ image pairs are collected from the Google Landmarks dataset \cite{noh2017large}. Local feature matching is performed via nearest neighbor search followed by geometric verification (RANSAC \cite{Fischler1981} with an affine model).
\figref{fig:plot_matching} plots the relevance probabilities as a function of the DELF local feature attention scores (these attention scores can be interpreted as a measure of a local feature's ``landmarkness''). The blue curve denotes features that are located within bounding boxes, while the red curve represents features located outside bounding boxes.

The curves show that local features located within bounding boxes are much more likely to be relevant: for two features with the same attention score, the relevance probability for a feature located within a predicted box is approximately $3$ to $4\times$ larger than that for a local feature located outside the box. Note how feature relevance increases with attention scores, as expected, but the predicted boxes can provide important extra information to effectively select the best features. This can be interpreted as the merging of two information streams: \textit{bottom-up} (DELF attention scores estimate per-local feature relevance) and \textit{top-down} (landmark detector estimates relevance of large regions).

Our proposed R-ASMK$^\star$ can be seen as a local feature re-weighting mechanism, which favors features located within detected regions.
The experimental results obtained on the $\mathcal{R}$Oxford and $\mathcal{R}$Paris datasets confirm that re-weighting features within detected regions boosts image retrieval performance substantially.

\section*{Appendix B. Detection Experiments}

We present learning curves for the trained detectors, and examples of detected regions compared to ground-truth.

\paragraph{Learning Curves.}

\begin{figure}[t]
\begin{center}
   \includegraphics[width=1\linewidth]{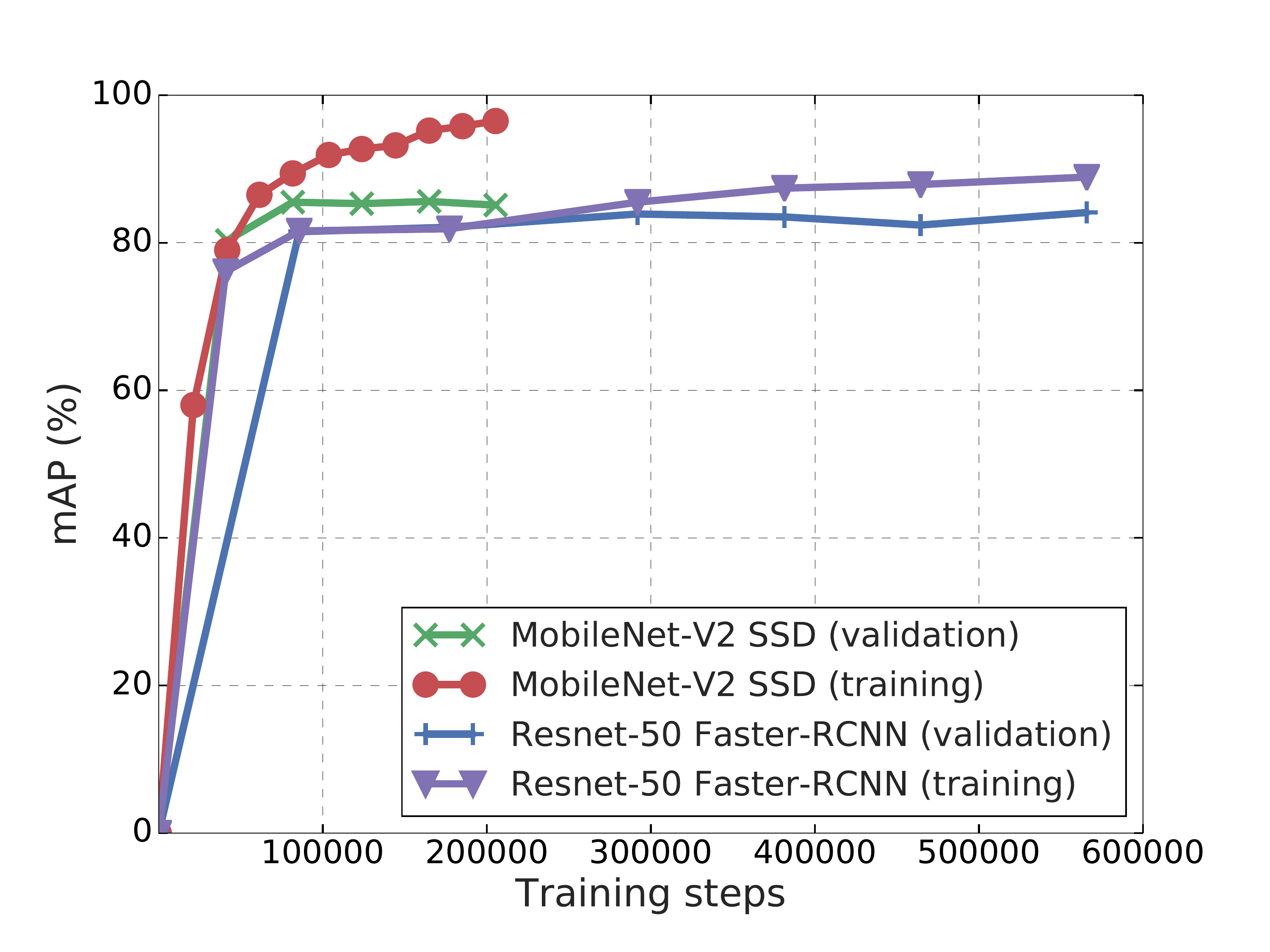}
\end{center}
\vspace{-15pt}
   \caption{Mean average precision @ IOU=$0.5$ for the two trained landmark detectors, as a function of the number of training steps.}
\label{fig:map_curves}
\vspace{0pt}      
\end{figure}

We train both Faster-RCNN and SSD based object detection models on our dataset. \figref{fig:map_curves} shows the comparison of learning progression of the two models. Both models converge to around 85\% mAP within 600k training steps. The MobilenetV2 SSD model trains much faster than the Resnet-50 Faster-RCNN, due to much smaller model size and larger batch size ($32$ vs. $1$, respectively). We also observe that SSD-based model slightly outperforms the Faster-RCNN base model despite having a smaller/weaker feature extractor. We conjecture that the advantage is due to the multi-scale feature map of SSD capturing the landmarks at different scales better than Faster-RCNN, which operates on a single feature map.

\paragraph{Detection Examples.}

\begin{figure*}[b]
\begin{center}
   \includegraphics[width=1\linewidth]{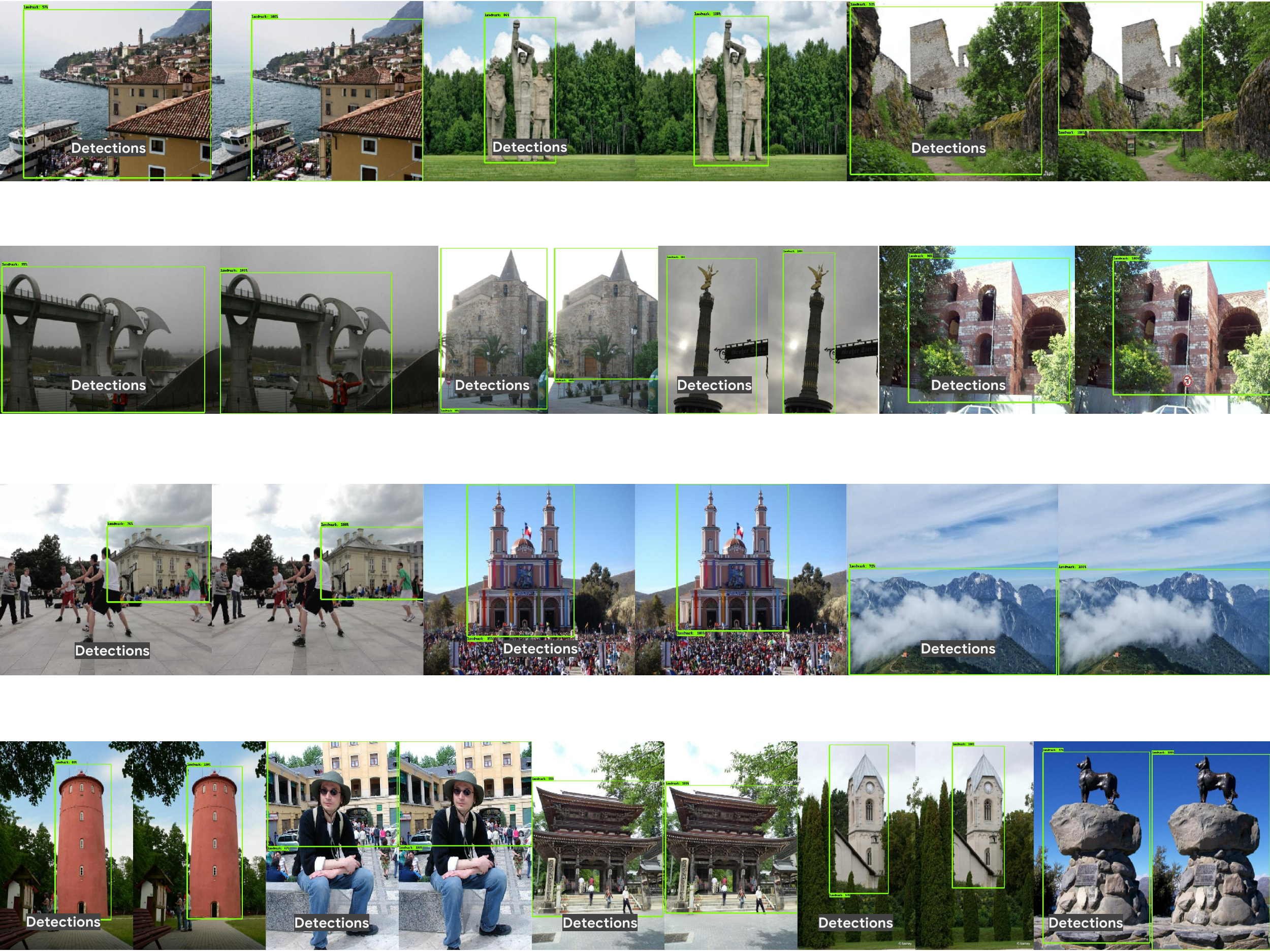}
\end{center}
\vspace{-15pt}
   \caption{Detection (on the left) versus ground-truth (on the right) on the Google Landmarks dataset.}
\label{fig:gld_detection_examples}
\vspace{25pt}      
\end{figure*}

\begin{figure*}[t]
\begin{center}
   \includegraphics[width=0.45\linewidth]{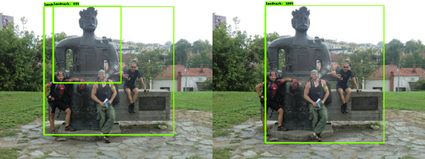}
   \includegraphics[width=0.45\linewidth]{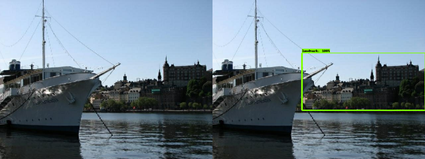}
\end{center}
\vspace{-15pt}
   \caption{Two failure detection cases. On the right are the ground-truth images, and on the left are the outputs of the detector (if any).}
\label{fig:gld_detection_failures_examples}
\vspace{0pt}      
\end{figure*}

To illustrate the effectiveness of our trained detectors, we present examples of detection using the SSD model. \figref{fig:gld_detection_examples} shows examples for a variety of landmarks with different scales, occlusion and lighting conditions. In addition, we also show some failure cases in \figref{fig:gld_detection_failures_examples} where the object of interest has ambiguous semantic boundary (resulting in double-detection) or is very hard to distinguish from the scene (resulting in missed detection).
For both figures, only detections with confidence probability more than $0.2$ are shown.

\section*{Appendix C. Region Selection Comparison}

\begin{figure*}
\begin{tabular}{ccc}

\includegraphics[height=0.2 \textwidth]{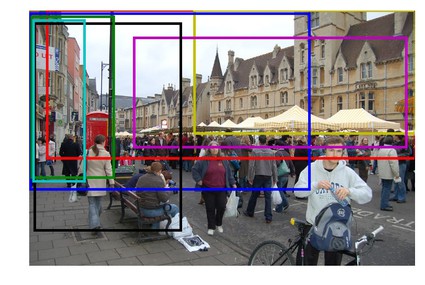}&
\includegraphics[height=0.2 \textwidth]{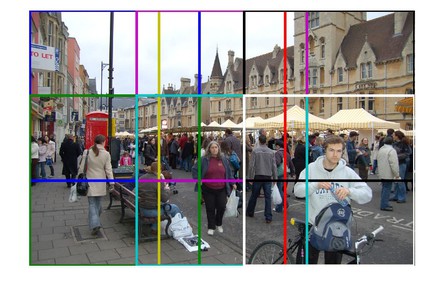}&
\includegraphics[height=0.2 \textwidth]{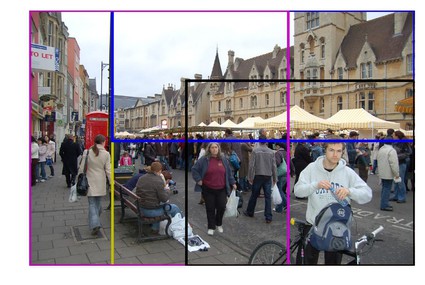} \\

\includegraphics[height=0.2 \textwidth]{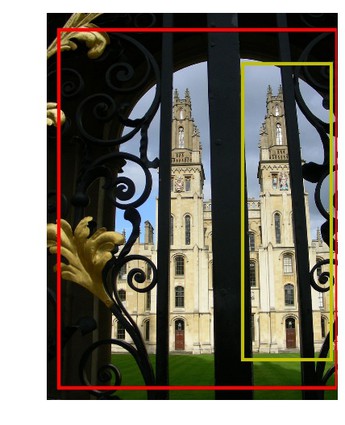}&
\includegraphics[height=0.2 \textwidth]{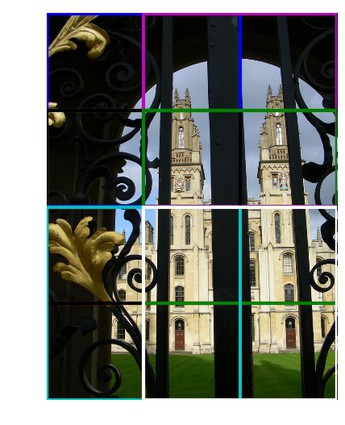}&
\includegraphics[height=0.2 \textwidth]{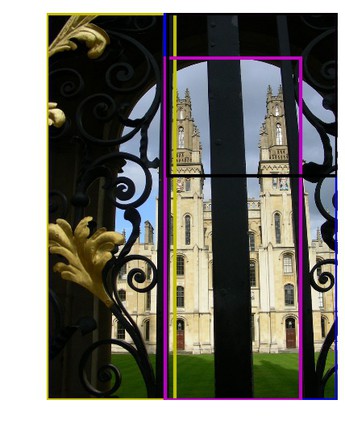} \\

\includegraphics[height=0.2 \textwidth]{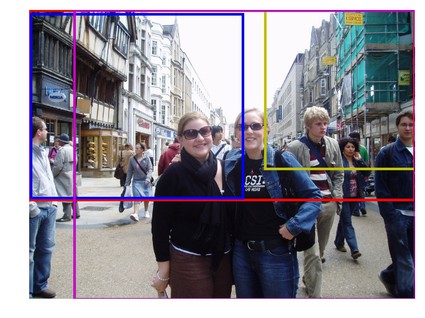}&
\includegraphics[height=0.2 \textwidth]{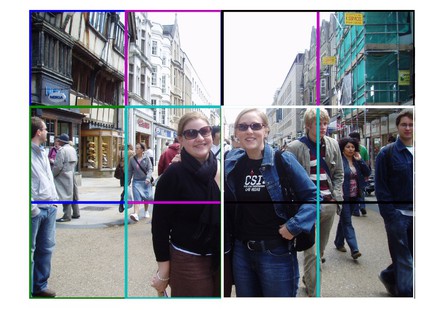}&
\includegraphics[height=0.2 \textwidth]{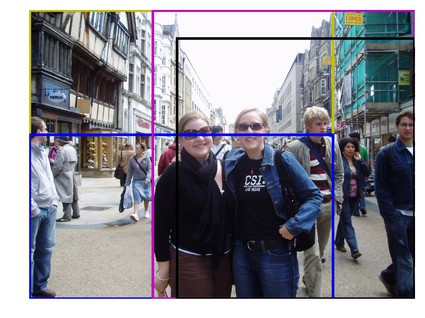} \\

\includegraphics[height=0.2 \textwidth]{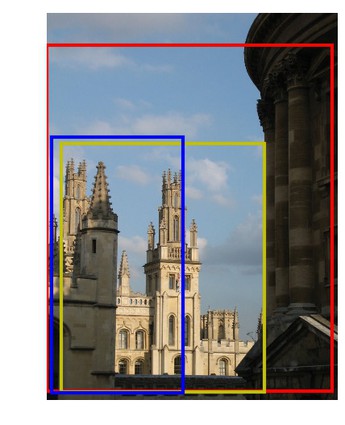}&
\includegraphics[height=0.2 \textwidth]{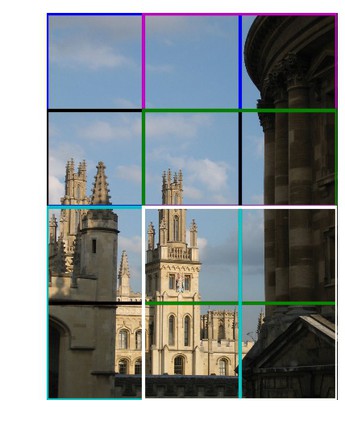}&
\includegraphics[height=0.2 \textwidth]{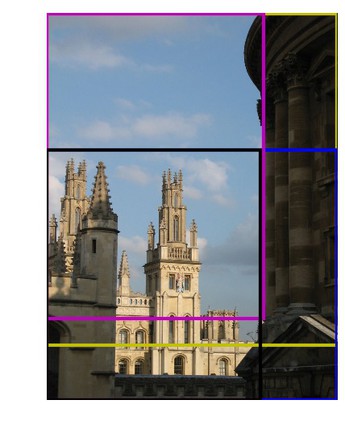} \\

\includegraphics[height=0.2 \textwidth]{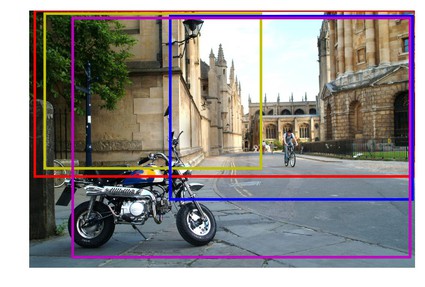}&
\includegraphics[height=0.2 \textwidth]{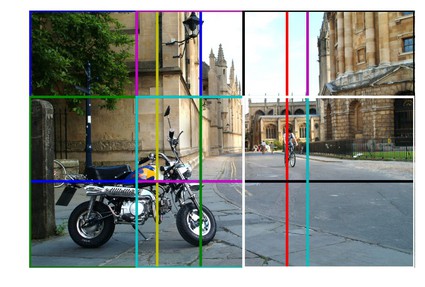}&
\includegraphics[height=0.2 \textwidth]{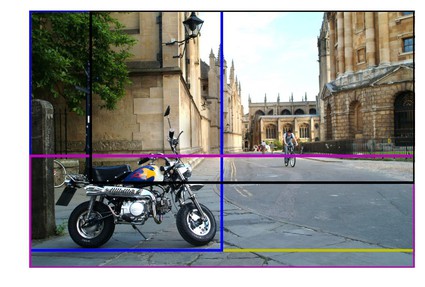} \\

\includegraphics[height=0.2 \textwidth]{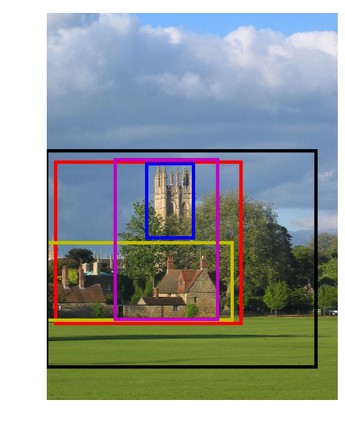}&
\includegraphics[height=0.2 \textwidth]{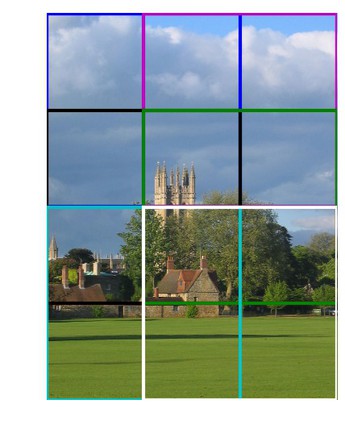}&
\includegraphics[height=0.2 \textwidth]{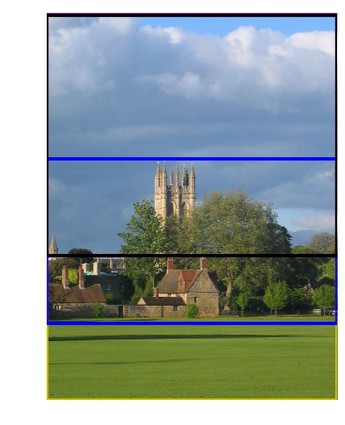}

\end{tabular}
\caption{Examples of selected regions for the three methods compared in the paper, on the $\mathcal{R}$Oxford dataset. Left: our D2R approach, with detection threshold of $0.1$ ($4.1$ regions per image). Center: RMAC boxes (fixed multi-scale grid), with 2 levels ($9$ regions per image). Right: Selective search, with $6$ regions per image. Note that edges for some regions overlap in some cases, so not all regions may be clearly visible.}
\label{fig:roxford_detection}
\end{figure*}

\begin{figure*}
\begin{tabular}{ccc}

\includegraphics[height=0.2 \textwidth]{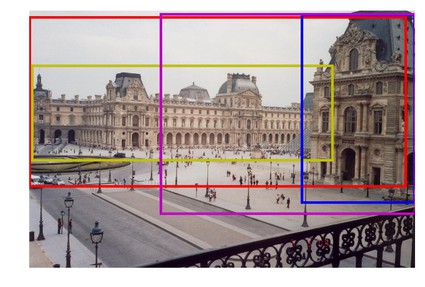}&
\includegraphics[height=0.2 \textwidth]{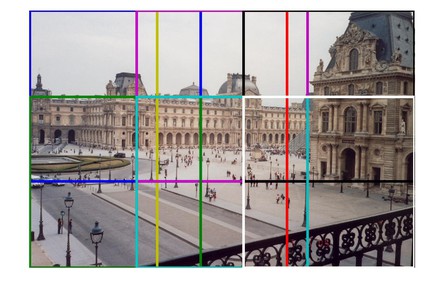}&
\includegraphics[height=0.2 \textwidth]{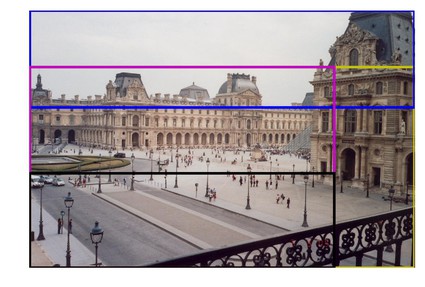} \\

\includegraphics[height=0.2 \textwidth]{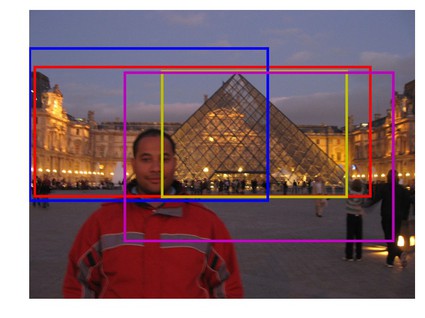}&
\includegraphics[height=0.2 \textwidth]{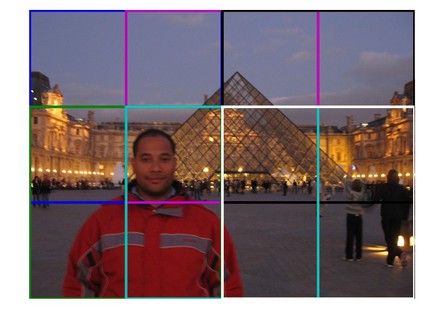}&
\includegraphics[height=0.2 \textwidth]{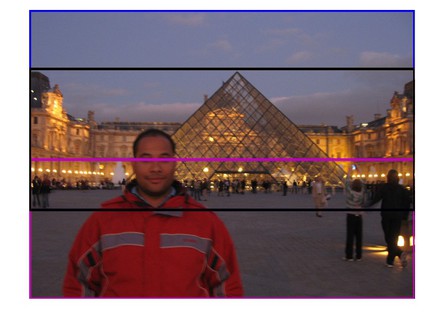} \\

\includegraphics[height=0.2 \textwidth]{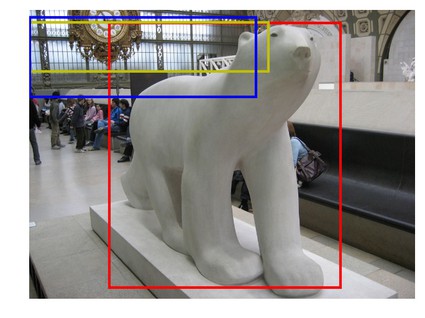}&
\includegraphics[height=0.2 \textwidth]{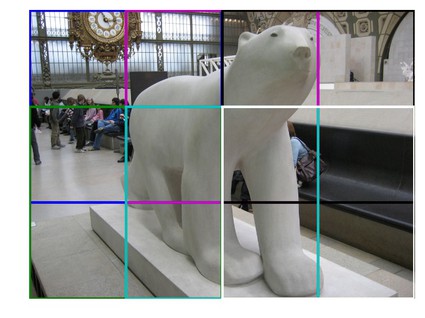}&
\includegraphics[height=0.2 \textwidth]{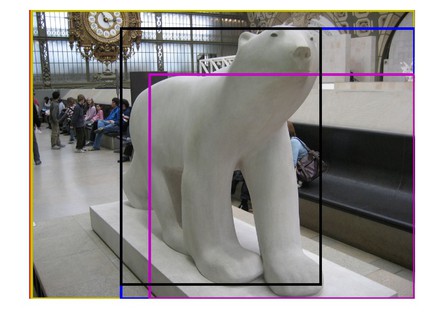} \\

\includegraphics[height=0.2 \textwidth]{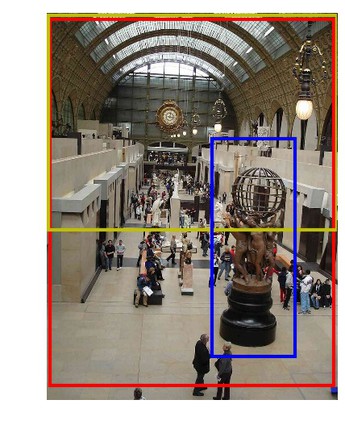}&
\includegraphics[height=0.2 \textwidth]{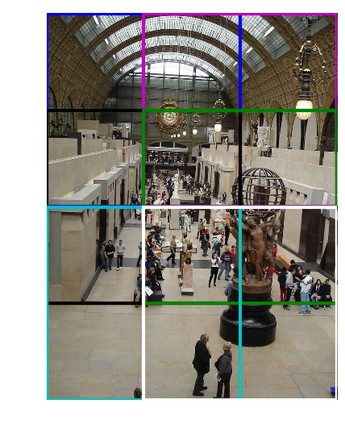}&
\includegraphics[height=0.2 \textwidth]{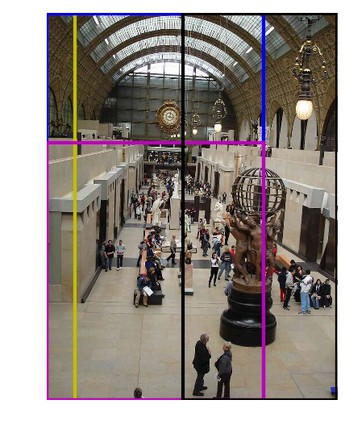} \\

\includegraphics[height=0.2 \textwidth]{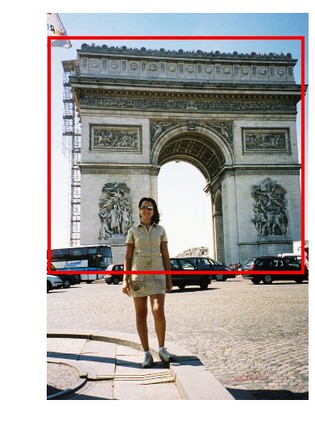}&
\includegraphics[height=0.2 \textwidth]{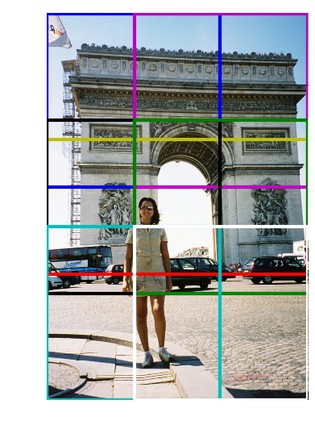}&
\includegraphics[height=0.2 \textwidth]{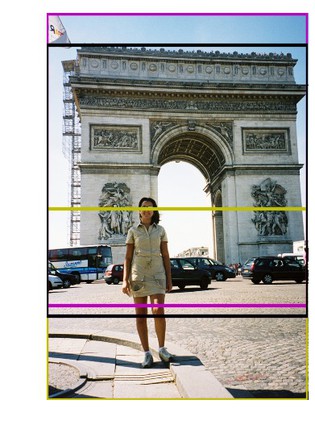} \\

\includegraphics[height=0.2 \textwidth]{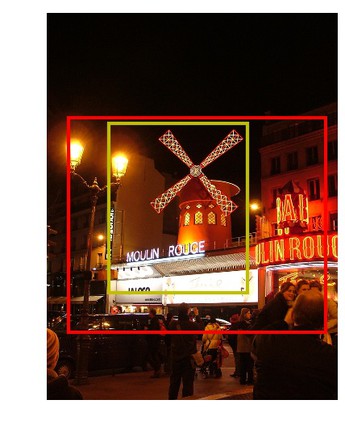}&
\includegraphics[height=0.2 \textwidth]{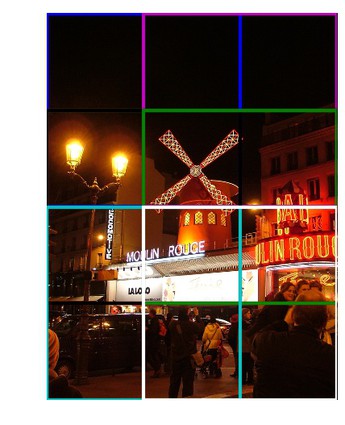}&
\includegraphics[height=0.2 \textwidth]{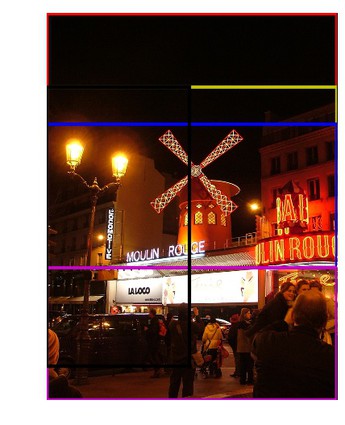}

\end{tabular}
\caption{Examples of selected regions for the three methods compared in the paper, on the $\mathcal{R}$Paris dataset. Left: our D2R approach, with detection threshold of $0.1$ ($3.9$ regions per image). Center: RMAC boxes (fixed multi-scale grid), with 2 levels ($9$ regions per image). Right: Selective search, with $6$ regions per image. Note that edges for some regions overlap in some cases, so not all regions may be clearly visible.}
\label{fig:rparis_detection}
\end{figure*}

In this section, we present landmark detection results on the $\mathcal{R}$Oxford and $\mathcal{R}$Paris datasets (\figref{fig:roxford_detection} and \figref{fig:rparis_detection}, respectively), comparing with the selected regions by competitive approaches (RMAC boxes and Selective Search).
The three methods use a configuration that produces a roughly similar number of regions per image: D2R with detection threshold $0.1$ (about $4$ regions per image), RMAC boxes with $2$ levels ($9$ regions per image), and Selective Search with $6$ selected regions per image. Note that our image retrieval experiments always use the whole image as a valid region, but in these visualizations we do not box the whole image, for a more concise presentation.

The figures show that our trained landmark detector tends to focus on the most prominent landmark regions in the image. RMAC boxes correspond to a fixed multi-scale grid, where the selected regions only depend on the input image size, not on its contents. This leads to regularly spaced boxes which do not usually overlap well with landmarks. Selective search produces boxes corresponding to prominent objects in the scene, which may or may not correspond to landmarks.

{\small
\bibliographystyle{ieee}
\bibliography{literature/delf}

\begin{thebibliography}{10}\itemsep=-1pt

\bibitem{arandjelovic2016netvlad}
R.~Arandjelovi{\'c}, P.~Gronat, A.~Torii, T.~Pajdla, and J.~Sivic.
\newblock {NetVLAD: CNN Architecture for Weakly Supervised Place Recognition}.
\newblock In {\em Proc. CVPR}, 2016.

\bibitem{arandjelovic2013all}
R.~Arandjelovic and A.~Zisserman.
\newblock {All About VLAD}.
\newblock In {\em Proc. CVPR}, 2013.

\bibitem{avrithis2014hough}
Y.~Avrithis and G.~Tolias.
\newblock {Hough Pyramid Matching: Speeded-up Geometry Re-ranking for Large
  Scale Image Retrieval}.
\newblock {\em IJCV}, 2014.

\bibitem{babenko2015iccv}
A.~Babenko and V.~Lempitsky.
\newblock {Aggregating Local Deep Features for Image Retrieval}.
\newblock In {\em Proc. ICCV}, 2015.

\bibitem{babenko2014neural}
A.~Babenko, A.~Slesarev, A.~Chigorin, and V.~Lempitsky.
\newblock {Neural Codes for Image Retrieval}.
\newblock In {\em Proc. ECCV}, 2014.

\bibitem{bay2008speeded}
H.~Bay, A.~Ess, T.~Tuytelaars, and L.~Van~Gool.
\newblock {Speeded-Up Robust Features (SURF)}.
\newblock {\em CVIU}, 2008.

\bibitem{chum2007total}
O.~Chum, J.~Philbin, J.~Sivic, M.~Isard, and A.~Zisserman.
\newblock {Total Recall: Automatic Query Expansion with a Generative Feature
  Model for Object Retrieval}.
\newblock In {\em Proc. ICCV}, 2007.

\bibitem{Fischler1981}
M.~Fischler and R.~Bolles.
\newblock {Random Sample Consensus: A Paradigm for Model Fitting with
  Applications to Image Analysis and Automated Cartography}.
\newblock {\em Communications of the ACM}, 1981.

\bibitem{gordo2016deep}
A.~Gordo, J.~Almazan, J.~Revaud, and D.~Larlus.
\newblock {Deep Image Retrieval: Learning Global Representations for Image
  Search}.
\newblock In {\em Proc. ECCV}, 2016.

\bibitem{he2015deep}
K.~He, X.~Zhang, S.~Ren, and J.~Sun.
\newblock {Deep Residual Learning for Image Recognition}.
\newblock In {\em Proc. CVPR}, 2016.

\bibitem{huang2017speed}
J.~Huang, V.~Rathod, C.~Sun, M.~Zhu, A.~Korattikara, A.~Fathi, I.~Fischer,
  Z.~Wojna, Y.~Song, S.~Guadarrama, et~al.
\newblock {Speed/Accuracy Trade-offs for Modern Convolutional Object
  Detectors}.
\newblock In {\em Proc. CVPR}, 2017.

\bibitem{iscen2018fast}
A.~Iscen, Y.~Avrithis, G.~Tolias, T.~Furon, and O.~Chum.
\newblock {Fast Spectral Ranking for Similarity Search}.
\newblock In {\em Proc. CVPR}, 2018.

\bibitem{iscen2017efficient}
A.~Iscen, G.~Tolias, Y.~Avrithis, T.~Furon, and O.~Chum.
\newblock {Efficient Diffusion on Region Manifolds: Recovering Small Objects
  with Compact CNN Representations}.
\newblock In {\em Proc. CVPR}, 2017.

\bibitem{Jegou2008}
H.~J\'{e}gou, M.~Douze, and C.~Schmid.
\newblock {Hamming Embedding and Weak Geometric Consistency for Large Scale
  Image Search}.
\newblock In {\em Proc. ECCV}, 2008.

\bibitem{jegou2010aggregating}
H.~J\'{e}gou, M.~Douze, C.~Schmidt, and P.~Perez.
\newblock {Aggregating Local Descriptors into a Compact Image Representation}.
\newblock In {\em Proc. CVPR}, 2010.

\bibitem{jegou2012aggregating}
H.~J\'{e}gou, F.~Perronnin, M.~Douze, J.~Sanchez, P.~Perez, and C.~Schmid.
\newblock {Aggregating Local Image Descriptors into Compact Codes}.
\newblock {\em IEEE Transactions on Pattern Analysis and Machine Intelligence},
  2012.

\bibitem{kim2015predicting}
H.~J. Kim, E.~Dunn, and J.-M. Frahm.
\newblock {Predicting Good Features for Image Geo-Localization Using Per-Bundle
  VLAD}.
\newblock In {\em Proc. ICCV}, 2015.

\bibitem{krizhevsky2012imagenet}
A.~Krizhevsky, I.~Sutskever, and G.~Hinton.
\newblock {Imagenet Classification with Deep Convolutional Neural Networks}.
\newblock In {\em Proc. NIPS}, 2012.

\bibitem{kuznetsova2018open}
A.~Kuznetsova, H.~Rom, N.~Alldrin, J.~Uijlings, I.~Krasin, J.~Pont-Tuset,
  S.~Kamali, S.~Popov, M.~Malloci, T.~Duerig, and V.~Ferrari.
\newblock {The Open Images Dataset V4: Unified Image Classification, Object
  Detection, and Visual Relationship Detection at Scale}.
\newblock {\em arXiv:1811.00982}, 2018.

\bibitem{liu2016ssd}
W.~Liu, D.~Anguelov, D.~Erhan, C.~Szegedy, S.~Reed, C.-Y. Fu, and A.~C. Berg.
\newblock {SSD: Single Shot Multibox Detector}.
\newblock In {\em Proc. ECCV}, 2016.

\bibitem{Lowe2004}
D.~Lowe.
\newblock {Distinctive Image Features from Scale-Invariant Keypoints}.
\newblock {\em IJCV}, 2004.

\bibitem{matas2004robust}
J.~Matas, O.~Chum, M.~Urban, and T.~Pajdla.
\newblock {Robust Wide-Baseline Stereo from Maximally Stable Extremal Regions}.
\newblock {\em Image and Vision Computing}, 2004.

\bibitem{mishchuk2017working}
A.~Mishchuk, D.~Mishkin, F.~Radenovic, and J.~Matas.
\newblock {Working Hard to Know your Neighbor's Margins: Local Descriptor
  Learning Loss}.
\newblock In {\em Proc. NIPS}, 2017.

\bibitem{mishkin2018repeatability}
D.~Mishkin, F.~Radenovic, and J.~Matas.
\newblock {Repeatability Is Not Enough: Learning Affine Regions via
  Discriminability}.
\newblock In {\em Proc. ECCV}, 2018.

\bibitem{noh2017large}
H.~Noh, A.~Araujo, J.~Sim, T.~Weyand, and B.~Han.
\newblock {Large-Scale Image Retrieval with Attentive Deep Local Features}.
\newblock In {\em Proc. ICCV}, 2017.

\bibitem{Philbin07}
J.~Philbin, O.~Chum, M.~Isard, J.~Sivic, and A.~Zisserman.
\newblock {Object Retrieval with Large Vocabularies and Fast Spatial Matching}.
\newblock In {\em Proc. CVPR}, 2007.

\bibitem{Philbin2008}
J.~Philbin, O.~Chum, M.~Isard, J.~Sivic, and A.~Zisserman.
\newblock {Lost in Quantization: Improving Particular Object Retrieval in Large
  Scale Image Databases}.
\newblock In {\em Proc. CVPR}, 2008.

\bibitem{radenovic2018revisiting}
F.~Radenovi{\'c}, A.~Iscen, G.~Tolias, Y.~Avrithis, and O.~Chum.
\newblock {Revisiting Oxford and Paris: Large-Scale Image Retrieval
  Benchmarking}.
\newblock In {\em Proc. CVPR}, 2018.

\bibitem{radenovic2016cnn}
F.~Radenovi{\'c}, G.~Tolias, and O.~Chum.
\newblock {CNN Image Retrieval Learns from BoW: Unsupervised Fine-Tuning with
  Hard Examples}.
\newblock In {\em Proc. ECCV}, 2016.

\bibitem{radenovic2018fine}
F.~Radenovi{\'c}, G.~Tolias, and O.~Chum.
\newblock {Fine-tuning CNN Image Retrieval with No Human Annotation}.
\newblock {\em IEEE Transactions on Pattern Analysis and Machine Intelligence},
  2018.

\bibitem{razavian2016visual}
A.~S. Razavian, J.~Sullivan, S.~Carlsson, and A.~Maki.
\newblock {Visual Instance Retrieval with Deep Convolutional Networks}.
\newblock {\em ITE Transactions on Media Technology and Applications}, 2016.

\bibitem{ren2015faster}
S.~Ren, K.~He, R.~Girshick, and J.~Sun.
\newblock {Faster R-CNN: Towards Real-Time Object Detection with Region
  Proposal Networks}.
\newblock In {\em Proc. NIPS}, 2015.

\bibitem{sandler2018inverted}
M.~Sandler, A.~Howard, M.~Zhu, A.~Zhmoginov, and L.-C. Chen.
\newblock {Inverted Residuals and Linear Bottlenecks: Mobile Networks for
  Classification, Detection and Segmentation}.
\newblock In {\em Proc. CVPR}, 2018.

\bibitem{simeoni2019local}
O.~Simeoni, Y.~Avrithis, and O.~Chum.
\newblock {Local Features and Visual Words Emerge in Activations}.
\newblock In {\em Proc. CVPR}, 2019.

\bibitem{Simonyan15}
K.~Simonyan and A.~Zisserman.
\newblock {Very Deep Convolutional Networks for Large-Scale Image Recognition}.
\newblock In {\em Proc. ICLR}, 2015.

\bibitem{Sivic2003}
J.~Sivic and A.~Zisserman.
\newblock {Video Google: A Text Retrieval Approach to Object Matching in
  Videos}.
\newblock In {\em Proc. ICCV}, 2003.

\bibitem{TaoCVPR2014}
R.~Tao, E.~Gavves, C.~G.~M. Snoek, and A.~W.~M. Smeulders.
\newblock {Locality in Generic Instance Search from One Example}.
\newblock In {\em Proc. CVPR}, 2014.

\bibitem{tolias2015image}
G.~Tolias, Y.~Avrithis, and H.~Jegou.
\newblock {Image Search with Selective Match Kernels: Aggregation Across Single
  and Multiple Images}.
\newblock {\em IJCV}, 2015.

\bibitem{tolias2014visual}
G.~Tolias and H.~Jegou.
\newblock {Visual Query Expansion with or without Geometry: Refining Local
  Descriptors by Feature Aggregation}.
\newblock {\em Pattern Recognition}, 2014.

\bibitem{tolias2015particular}
G.~Tolias, R.~Sicre, and H.~J{\'e}gou.
\newblock {Particular Object Retrieval with Integral Max-Pooling of CNN
  Activations}.
\newblock In {\em Proc. ICLR}, 2015.

\bibitem{UijlingsIJCV2013}
J.~R.~R. Uijlings, K.~E.~A. van~de Sande, T.~Gevers, and A.~W.~M. Smeulders.
\newblock {Selective Search for Object Recognition}.
\newblock {\em IJCV}, 2013.

\end{thebibliography}
}

\end{document}